%% file: main.tex
\documentclass[10pt,twocolumn,letterpaper]{article}

\usepackage{wacv}              % To produce the CAMERA-READY version

\usepackage{booktabs}
\usepackage{hhline}
\usepackage{url}
\usepackage{multirow}
\usepackage{arydshln}
\usepackage{graphicx}
\usepackage{subcaption}
\usepackage{tikz}
\usepackage{amssymb} 
\usepackage{amsmath}

\newcommand{\rev}[1]{\textcolor{black}{#1}}
\usepackage[accsupp]{axessibility}  % Improves PDF readability for those with disabilities.


\definecolor{wacvblue}{rgb}{0.21,0.49,0.74}
\usepackage[pagebackref,breaklinks,colorlinks,allcolors=wacvblue]{hyperref}

\def\wacvPaperID{1445} % *** Enter the WACV Paper ID here
\def\confName{WACV}
\def\confYear{2027}

\title{Hard Sample-Conditioned Decorrelated Attention for Robust Face Recognition}

\author {
    Tahar Chettaoui\textsuperscript{\textrm 1},
    Guray Ozgur\textsuperscript{\textrm 1, \textrm 2},
    Eduarda Caldeira\textsuperscript{\textrm 1, \textrm 2},
    Naser Damer\textsuperscript{\textrm 1, \textrm 2},
    Fadi Boutros\textsuperscript{\textrm 1}\\
    \textsuperscript{\rm 1}Fraunhofer Institute for Computer Graphics Research IGD, Germany\\
    \textsuperscript{\rm 2}TU Darmstadt, Germany
    \\
    {\tt\small tahar.chettaoui@igd.fraunhofer.de}
}

\begin{document}
\maketitle
\input{sec/0_abstract}    
\input{sec/1_intro}
\input{sec/2_related_work}

\input{sec/3_methodology}

\input{sec/4_experimnetal_setup}
\input{sec/5_results}
\input{sec/6_conclusion}
{
    \small
    \bibliographystyle{ieeenat_fullname}
    \bibliography{main}
}

\end{document}

%% file: sec/0_abstract.tex
\begin{abstract}

Multi-head attention in Vision Transformers (ViTs) allows independent attention patterns across heads, but without explicit regularization, heads may learn redundant patterns rather than complementary roles. Correlation-based penalties encourage head disentanglement, yet apply uniform regularization across all training samples. In face recognition (FR), hard samples whose embeddings lie close to competing identities are critical for optimization and generalization. Motivated by this, we propose ViT-HarDis, which conditions attention head disentanglement on sample difficulty: samples with high cosine similarity to negative class centers receive stronger head decorrelation, enforcing specialization where identity ambiguity is highest. Experiments show improvements over both the baseline and uniform disentanglement on most benchmarks. Our analysis shows that ViT-HarDis reduces the similarity of hard samples to competing class centers while decreasing inter-head attention similarity, and that applying disentanglement only to the final transformer layers outperforms regularizing the entire network. Trained on WebFace4M, ViT-HarDis achieves SOTA results among ViT-based methods on IJB-B, IJB-C, and TinyFace Rank-1, while achieving the best IJB-B and TinyFace results when trained on MS1MV3, improving FR without architectural modifications or additional inference cost. Code: \url{https://github.com/TaharChettaoui/Hard-Sample-Conditioned-Attention-for-FR}.
\end{abstract}

%% file: sec/1_intro.tex
\vspace{-5mm}
\section{Introduction}\label{sec:intro}

Face recognition (FR) has achieved remarkable progress in recent years, driven by advances in deep neural network architectures, innovations in training objectives, particularly margin-based softmax losses, and the availability of large-scale annotated face datasets \cite{DBLP:conf/cvpr/DengGXZ19, DBLP:conf/iccvw/DengGZDLS19, DBLP:conf/cvpr/Kim0L22, DBLP:conf/cvpr/BoutrosDKK22, wang2018cosfacelargemargincosine}. Recent years have witnessed a shift from convolutional neural networks (CNNs) \cite{DBLP:conf/cvpr/HeZRS16} to Vision Transformers (ViTs) \cite{dosovitskiy2021imageworth16x16words}, which leverage multi-head attention to capture complex spatial relationships and identity-discriminative patterns. ViT-based FR systems have demonstrated competitive performance on evaluation benchmarks, establishing themselves as a promising direction for FR \cite{DBLP:journals/ivc/ChettaouiDB25, DBLP:conf/fgr/ChettaouiOCDB26, DBLP:conf/cvpr/KimS0JL24, DBLP:conf/bmvc/SunT22, DBLP:conf/iccv/DanLXD0XS23, DBLP:conf/fgr/ChettaouiOCDB26a}.

Although multi-head attention enables ViT to capture diverse representations, prior works \cite{voita2019analyzing, michel2019sixteen, li2023attention} in general computer vision have shown that many attention heads become highly correlated or redundant, thus limiting the effective utilization of model capacity. \rev{An} existing attention head disentanglement method \cite{DBLP:conf/cvpr/BhattacharyyaCN23}, primarily developed for general vision tasks, \rev{addresses} this issue by uniformly penalizing correlations between attention heads. However, this approach \rev{treats} all samples equally and \rev{does} not account for differences in sample difficulty. In practice, hard samples that lie near decision boundaries are more critical for learning and generalization \cite{DBLP:conf/cvpr/HuangWT0SLLH20, DBLP:conf/cvpr/MengZH021, DBLP:conf/cvpr/Kim0L22}. We argue that attention head specialization should be adaptive, applying stronger disentanglement to these challenging cases, better exploiting model capacity while preserving complementary attention patterns where they are most needed.

\vspace{-1mm}

Motivated by these observations, we propose ViT-HarDis, a hard sample-conditioned attention head disentanglement framework for ViT-based FR. Instead of uniformly encouraging attention head diversity, our method adaptively strengthens decorrelation according to sample difficulty, promoting specialization for identity-ambiguous samples while preserving the original architecture. Beyond this adaptive regularization strategy, we analyze attention head redundancy in ViT-based FR, revealing that increasing the number of attention heads leads to higher redundancy without improving recognition accuracy. Our experiments demonstrate that ViT-HarDis outperforms both the baseline ViT and uniform disentanglement (ViT-Dis) on most considered benchmarks, including IJB-B \cite{inproceedingsijbb}, IJB-C \cite{DBLP:conf/icb/MazeADKMO0NACG18}, and TinyFace \cite{DBLP:conf/accv/ChengZG18}, while reducing the similarity of hard samples to competing identity centers and decreasing inter-head attention similarity. We further show that applying disentanglement only to the final transformer layers outperforms regularizing all layers, indicating that head specialization is layer-dependent. Finally, we train ViT-HarDis on MS1MV3 \cite{DBLP:conf/iccvw/DengGZDLS19} and WebFace4M \cite{zhu2021webface260mbenchmarkunveilingpower}, achieving SOTA results among ViT-based methods on IJB-B, IJB-C, and TinyFace Rank-1 when trained on WebFace4M, and the best IJB-B and TinyFace results when trained on MS1MV3, without architectural modifications or additional inference cost.

%Motivated by these observations, we propose ViT-HarDis, a hard sample-conditioned attention head disentanglement framework for ViT-based FR. Instead of uniformly encouraging head diversity, our method adaptively strengthens decorrelation according to sample difficulty, promoting specialization for identity-ambiguous samples while preserving the original architecture. We further provide a systematic analysis of attention head redundancy in ViT-based FR, revealing that increasing the number of heads leads to higher redundancy without improving recognition accuracy. Our experiments demonstrate that ViT-HarDis consistently outperforms both the baseline ViT and uniform disentanglement (ViT-Dis) on IJB-B \cite{inproceedingsijbb}, IJB-C \cite{DBLP:conf/icb/MazeADKMO0NACG18}, and TinyFace \cite{DBLP:conf/accv/ChengZG18}, while reducing the similarity of hard samples to competing identity centers and decreasing inter-head attention similarity. We also show that applying disentanglement only to the final transformer layers outperforms regularizing all layers, indicating that head specialization is layer-dependent. Finally, we train ViT-HarDis on MS1MV3 \cite{DBLP:conf/iccvw/DengGZDLS19} and WebFace4M \cite{zhu2021webface260mbenchmarkunveilingpower}, achieving SOTA results among ViT-based methods on IJB-B, IJB-C, and TinyFace Rank-1 with WebFace4M, while remaining competitive with MS1MV3, without architectural modifications or additional inference cost.

%% file: sec/2_related_work.tex
\section{Related Work}
\label{sec:related_work}

\subsection{Vision Transformers for Face Recognition}
ViTs~\cite{dosovitskiy2021imageworth16x16words} have recently emerged as a competitive alternative to CNNs for FR, owing to their ability to capture long-range spatial dependencies through self-attention. Early studies demonstrated the feasibility of applying ViTs to FR by learning discriminative facial representations from local facial regions. For example, Sun et al. \cite{DBLP:conf/bmvc/SunT22} proposed a landmark-guided part-based ViT that aggregates local facial patches to improve recognition performance, while ARTriViT~\cite{10228106} introduced a Siamese ViT architecture trained with triplet loss for metric learning.
Subsequent works focused on improving the robustness and scalability of ViT-based FR models. TransFace~\cite{DBLP:conf/iccv/DanLXD0XS23} proposed dominant patch perturbation and entropy-based hard sample mining to improve optimization and generalization. KP-RPE~\cite{DBLP:conf/cvpr/KimS0JL24} introduced keypoint-relative positional encoding to enhance robustness against pose and geometric variations by incorporating facial landmark information into the attention mechanism. More recently, FRoundation~\cite{DBLP:journals/ivc/ChettaouiDB25} demonstrated that large pre-trained ViT-based foundation models can be effectively adapted to FR using parameter-efficient fine-tuning, while SwinFace~\cite{DBLP:journals/tcsv/QinWDWCHD24} extended transformer-based facial analysis through a multi-task learning framework that jointly performs FR, expression recognition, age estimation, and facial attribute prediction. Chettaoui et al. \cite{DBLP:conf/fgr/ChettaouiOCDB26} further improved ViT-based FR by introducing register tokens to better structure attention and reduce attention artifacts.

%Despite these advances, existing ViT-based FR methods primarily focus on improving network architectures, positional encoding, or training strategies \cite{DBLP:conf/cvpr/KimS0JL24,DBLP:conf/iccv/DanLXD0XS23}, while the behavior of individual attention heads remains largely unexplored. In particular, they do not explicitly encourage different attention heads to learn complementary identity-discriminative representations.

Despite these advances, existing ViT-based FR methods focus on improving network architectures, positional encoding, or training strategies~\cite{DBLP:conf/cvpr/KimS0JL24,DBLP:conf/iccv/DanLXD0XS23}, while the behavior of individual attention heads remains largely unexplored. In particular, they do not explicitly encourage heads to learn complementary identity-discriminative features.

\subsection{Attention Head Disentanglement}

Multi-head self-attention enables ViTs to capture complementary information by attending to multiple representation subspaces~\cite{dosovitskiy2021imageworth16x16words}. Nevertheless, several studies have shown that many attention heads become redundant after training \cite{michel2019sixteen, voita2019analyzing, li2023attention}. Voita \textit{et al.}~\cite{voita2019analyzing} demonstrated that only a small subset of attention heads contributes substantially to model performance, while many remaining heads can be pruned with negligible performance degradation. Similar observations have also been reported for ViT, where heads often exhibit highly correlated attention patterns and unequal contributions to the final prediction~\cite{li2023attention}. These findings suggest that multi-head attention capacity is often underutilized.

%Similar observations have also been reported for ViT, where heads often exhibit highly correlated patterns and unequal contributions to the final prediction~\cite{...}.

Motivated by these observations, DeCAtt~\cite{DBLP:conf/cvpr/BhattacharyyaCN23} introduces a correlation-based regularization that penalizes similarities between attention maps of different heads, encouraging complementary attention patterns without modifying the architecture. \rev{DeCAtt was evaluated on general vision tasks rather than FR, and applies the same regularization across all training samples, irrespective of their difficulty.}

Meanwhile, hard sample learning has become an essential component of modern FR systems. Methods such as CurricularFace~\cite{DBLP:conf/cvpr/HuangWT0SLLH20}, MagFace~\cite{DBLP:conf/cvpr/MengZH021} and AdaFace~\cite{DBLP:conf/cvpr/Kim0L22} demonstrate that emphasizing hard samples improves discriminative representation learning. Unlike these methods, which incorporate sample difficulty into the optimization objective, our approach leverages sample difficulty to adaptively guide attention head disentanglement. By enforcing stronger attention decorrelation particularly for identity-ambiguous samples, our method encourages more effective attention head specialization while preserving the original ViT architecture \cite{dosovitskiy2021imageworth16x16words}. To the best of our knowledge, this is the first work to condition attention head disentanglement on sample difficulty for ViT-based FR.

%% file: sec/3_methodology.tex
%%%%%%%%%%%%%%%%%%%%%%%%%%%%%%%%%%%%%%%%%%%%%%%%%%%%%%%%%%%%%%%%%%%%%%%%%%%%%%%%%%%%%%%%%%%%%%%%%%
\begin{figure}[!tp]
  \centering
   \includegraphics[width=1\linewidth]{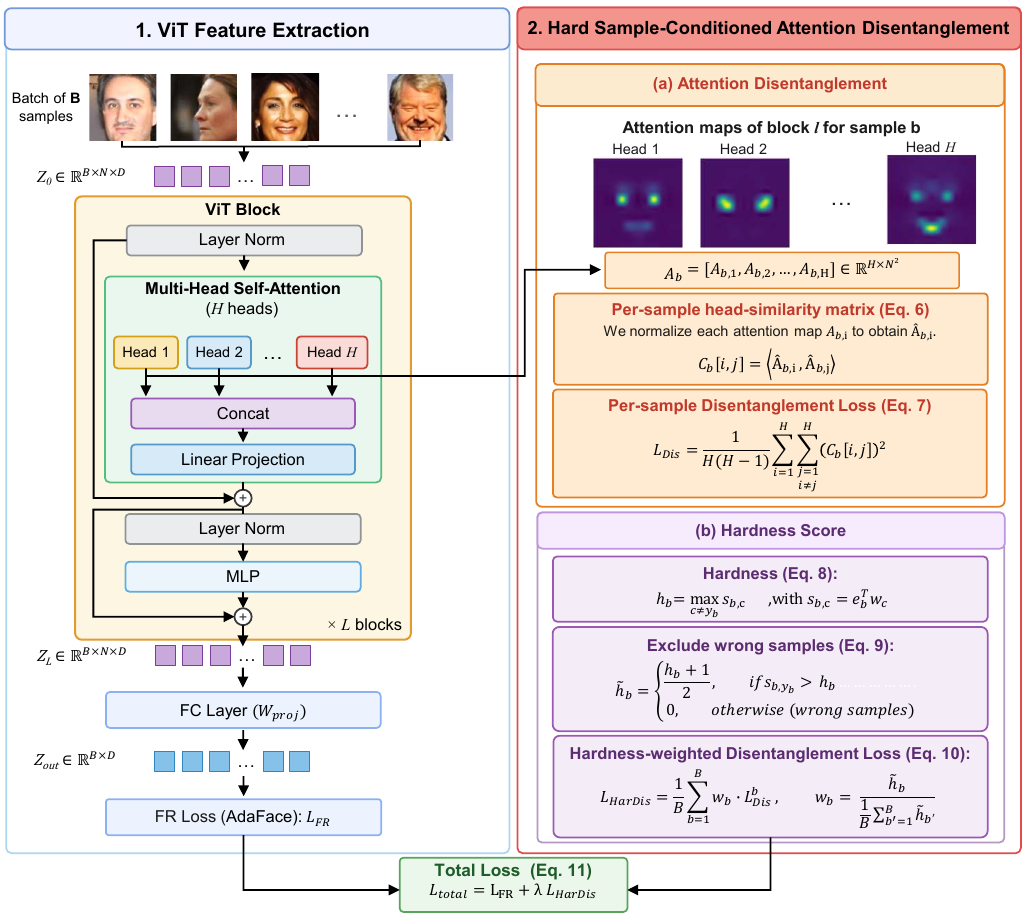}
\vspace{-5mm}
\caption{\rev{Overview of ViT-HarDis. A ViT with $L$ blocks and $H$ heads encodes a batch of $B$ face images (1). Per sample, the head-similarity matrix $\mathbf{C}_b$ of block $\ell$ gives the disentanglement loss $\mathcal{L}_{\text{Dis}}^{(b)}$ (2a), weighted by the sample hardness $w_b$ (2b). The weighted penalty $\mathcal{L}_{\text{HarDis}}$ is summed over the selected blocks and added to the FR loss with weight $\lambda$.}}
\label{fig:overview}
\vspace{-5mm}
\end{figure}

\vspace{-1mm}
\section{Methodology} \label{sec:method}
\vspace{-1mm}
%\rev{This section presents ViT-HarDis, our hard sample-conditioned approach to attention head disentanglement for ViT-based face recognition, illustrated in Figure \ref{fig:overview}. Attention head decorrelation reduces redundancy between heads, but applies a uniform penalty across all samples, independent of how difficult each is to recognize. However, samples differ in difficulty. Hard samples lie close to competing class centers and represent a challenge for recognition, whereas easy samples lie near their own class center. In ViT-HarDis, this penalty is instead conditioned on sample difficulty, quantified by the proximity of a sample's embedding to competing class centers. We first provide preliminaries on ViT and attention head disentanglement, followed by our proposed ViT-HarDis.}
\rev{This section presents ViT-HarDis, our hard sample-conditioned approach to attention head disentanglement for ViT-based FR, illustrated in Figure \ref{fig:overview}. Attention head decorrelation reduces redundancy between heads, but applies a uniform penalty across all samples, independent of how difficult each is to recognize. ViT-HarDis instead conditions this penalty on sample difficulty, quantified by the proximity of a sample's embedding to competing class centers, so that heads specialize where identity ambiguity is highest. We first provide preliminaries on ViT and attention head disentanglement, followed by our proposed ViT-HarDis.}

%%%%%%%%%%%%%%%%%%%%%%%%%%%%%%%%%%%%%%%%%%%%%%%%%%%%%%%%%%%%%%%%%%%%%%%%%%%%%%%%%%%%%%%%%%%%%%%%%%
\subsection{ViT (preliminary)} \label{sec:VIT}
Given a facial image \( X \in \mathbb{R}^{H_{in} \times W_{in} \times C_{in}} \), where \( H_{in} \), \( W_{in} \), and \( C_{in} \) represent the height, width, and number of channels, respectively, the image is divided into a sequence of patches, each of size \( P \times P \). These patches are then flattened into vectors, resulting in a patch matrix \( \mathbf{x}_p \in \mathbb{R}^{N \times (P^2 C_{in})} \), where \( N = \frac{H_{in}W_{in}}{P^2} \) is the total number of patches. These patch vectors are then linearly projected into a latent embedding space of dimension \( D \) using a trainable projection matrix \( \mathbf{E} \in \mathbb{R}^{(P^2C_{in}) \times D} \):

%\begin{equation}
%\mathbf{z}_0^i = \mathbf{x}_p^i \mathbf{E}, \quad \text{for } i = 1, \dots, N.
%\end{equation}
\vspace{-2mm}
\begin{equation}
\rev{\mathbf{Z}_0 = \mathbf{x}_p \mathbf{E} \in \mathbb{R}^{N \times D}}.
\end{equation}
\vspace{-3mm}

%To retain positional information, learnable positional embeddings \( \rev{\mathbf{E}_{\text{pos}} \in \mathbb{R}^{N \times D}} \) are added to the input embeddings \( \mathbf{Z}_0 \):
To retain positional information, learnable positional embeddings \( \rev{\mathbf{E}_{\text{pos}} \in \mathbb{R}^{N \times D}} \) are added to \( \mathbf{Z}_0 \):

%\begin{equation}
%\mathbf{Z}_0 = [\mathbf{z}_0^1; \dots; \mathbf{z}_0^N], 
%\end{equation}
\vspace{-2mm}
\begin{equation}
\mathbf{Z}_0' = \mathbf{Z}_0 + \mathbf{E}_{\text{pos}}.
\end{equation}
\vspace{-3mm}

The resulting token matrix \( \mathbf{Z}_0' \) is then fed into the transformer encoder, which maintains the constant latent dimension $D$ throughout all layers. The ViT follows the original transformer encoder design \cite{DBLP:journals/corr/VaswaniSPUJGKP17}, consisting of alternating layers of \rev{Multi-head} Self-Attention (MSA) and Multilayer Perceptron (MLP) blocks, each preceded by Layer Normalization (LN) and followed by residual connections. 
In MSA, \rev{$H$} heads are run in parallel, each with its own set of \rev{query}, \rev{key}, and \rev{value}. For an input token matrix \rev{$\mathbf{x} \in \mathbb{R}^{N \times D}$}, these are computed in attention head $i$ as:

\vspace{-3mm}
\begin{equation}
Q_i = \mathbf{x} W_i^Q, \quad K_i = \mathbf{x} W_i^K, \quad V_i = \mathbf{x} W_i^V,
\end{equation}
\vspace{-3mm}

where \( W_i^Q, W_i^K, W_i^V \in \mathbb{R}^{D \times d} \) are learnable weight matrices, and $d = D/H$ represents the dimensionality of each attention head\rev{, so that $Q_i, K_i, V_i \in \mathbb{R}^{N \times d}$}. Each head computes scaled dot-product attention:

\vspace{-2mm}
\begin{equation}
\text{Attention}(Q\rev{_i}, K\rev{_i}, V\rev{_i})  = \mathrm{softmax}\left( \frac{Q\rev{_i} K\rev{_i}^\top}{\sqrt{d}} \right)V\rev{_i}.
\end{equation}
\vspace{-3mm}

The outputs of all heads are concatenated and projected through a linear transformation to form the final attention output. This allows the model to jointly attend to information from different representation subspaces, enhancing its ability to capture complex relationships across patches \cite{dosovitskiy2021imageworth16x16words}. The output is then passed to the MLP, completing the execution of a single transformer block. These repeated attention and MLP operations refine the token representations throughout the encoder depth, ultimately producing contextualized outputs used for downstream tasks \cite{dosovitskiy2021imageworth16x16words}.

%%%%%%%%%%%%%%%%%%%%%%%%%%%%%%%%%%%%%%%%%%%%%%%%%%%%%%%%%%%%%%%%%%%%%%%%%%%%%%%%%%%%%%%%%%%%%%%%%%
\textbf{ViT for FR.} \label{sec:vit-face}  
%Given the contextualized patch embeddings from the final transformer layer $\mathbf{Z}_L \in \mathbb{R}^{N \times D}$, we \rev{flatten them into a single vector $\mathbf{Z}_L^{\text{flat}} \in \mathbb{R}^{ND}$} and project it through a learnable projection layer $\mathbf{W}_{\text{proj}} \in \mathbb{R}^{ND \times D}$ to obtain the face representation:
\rev{After $L$ blocks of alternating MSA and MLP with residual connections, the} final tokens $\mathbf{Z}_L \in \mathbb{R}^{N \times D}$ are \rev{flattened into $\mathbf{Z}_L^{\text{flat}} \in \mathbb{R}^{ND}$} and projected through a learnable layer $\mathbf{W}_{\text{proj}} \in \mathbb{R}^{ND \times D}$ to obtain the face representation:

\vspace{-2mm}
\begin{equation}
\mathbf{Z}_{\text{out}} = \rev{\mathbf{Z}_L^{\text{flat}}}\, \mathbf{W}_{\text{proj}}, \quad \mathbf{Z}_{\text{out}} \in \mathbb{R}^{D}.
\end{equation}
\vspace{-3mm}

%To optimize our model for FR, we use a margin-penalty softmax loss. Specifically, we extend the architecture with an additional multi-class classification layer and utilize AdaFace \cite{DBLP:conf/cvpr/Kim0L22} loss as the margin-penalty softmax objective. AdaFace adaptively adjusts the angular margin based on feature magnitude, providing tighter constraints for high-confidence samples while allowing flexibility for uncertain predictions. %This adaptive mechanism stabilizes training and improves convergence for FR tasks.
To optimize our model for FR, we extend the architecture with a multi-class classification layer and use AdaFace \cite{DBLP:conf/cvpr/Kim0L22} as the margin-penalty softmax objective. AdaFace adaptively adjusts the angular margin based on feature magnitude, providing tighter constraints for high-confidence samples while allowing flexibility for uncertain predictions.

%%%%%%%%%%%%%%%%%%%%%%%%%%%%%%%%%%%%%%%%%%%%%%%%%%%%%%%%%%%%%%%%%%%%%%%%%%%%%%%%%%%%%%%%%%%%%%%%%%
\subsection{Attention Head Disentanglement} \label{sec:head-disentang}
The \rev{multi-head} attention mechanism allows independent attention patterns across $H$ heads, but without explicit regularization, different heads may learn redundant patterns, as shown in Section \ref{sec:res_heads1}. Bhattacharyya et al. \cite{DBLP:conf/cvpr/BhattacharyyaCN23} \rev{address} this by encouraging disentanglement among heads through a correlation-based penalty applied to the attention map.

%\rev{We follow DeCAtt's \cite{DBLP:conf/cvpr/BhattacharyyaCN23} per-head normalization and correlation penalty, but compute the correlation per sample rather than over the batch, which enables the hardness weighting in Section~\ref{sec:hard-dis}. 
\rev{We follow DeCAtt's \cite{DBLP:conf/cvpr/BhattacharyyaCN23} per-head normalization and correlation penalty, written as a batch average of per-sample terms, which exposes the per-sample quantity that our hardness weighting in Section~\ref{sec:hard-dis} modulates.}
Let $\mathbf{A} \in \mathbb{R}^{B \times H \times N^2}$ denote the tensor of flattened attention maps across a batch of size $B$, where $H$ is the number of heads and $N$ is the number of patches. \rev{Each $\mathbf{A}_{b,i} \in \mathbb{R}^{N^2}$ is the $N \times N$ post-softmax attention map of head $i$ for sample $b$, flattened into a vector.} We L2-normalize each attention map $\mathbf{A}_{b,i}$ to obtain $\hat{\mathbf{A}}_{b,i}$, and form the per-sample head similarity matrix $\mathbf{C}_b$, whose entries are the cosine similarities between the attention maps of different heads:

%\begin{equation}
%\rev{\mathbf{C}_b[i,j] = \hat{\mathbf{A}}_{b,i}^\top \hat{\mathbf{A}}_{b,j}, \qquad \mathbf{C}_b \in \mathbb{R}^{h \times h}.}
%\end{equation}

\vspace{-2mm}
\begin{equation}
\rev{\mathbf{C}_b[i,j] = \langle \hat{\mathbf{A}}_{b,i}, \hat{\mathbf{A}}_{b,j} \rangle, \qquad \mathbf{C}_b \in \mathbb{R}^{H \times H}.}
\end{equation}
\vspace{-3mm}

%\rev{The per-sample disentanglement loss averages the squared off-diagonal entries of $\mathbf{C}_b$, penalizing overlap between the attention maps of different heads while ignoring the diagonal:}
\rev{The per-sample disentanglement loss averages the squared off-diagonal entries of $\mathbf{C}_b$, penalizing overlap between the attention maps of different heads:}

\vspace{-2mm}
\begin{equation}
\rev{\mathcal{L}_{\text{Dis}}^{(b)} = \frac{1}{H(H-1)} \sum_{i=1}^{H} \sum_{\substack{j=1 \\ j \neq i}}^{H} \left( \mathbf{C}_b[i,j] \right)^2.}
\end{equation}
\vspace{-3mm}

%\rev{Finally, we average the per-sample disentanglement losses over the batch to obtain the uniform disentanglement objective $\mathcal{L}_{\text{Dis}} = \frac{1}{B}\sum_{b=1}^{B} \mathcal{L}_{\text{Dis}}^{(b)}$.}

Averaging over the batch gives the uniform disentanglement objective $\mathcal{L}_{\text{Dis}} = \frac{1}{B}\sum_{b=1}^{B} \mathcal{L}^{(b)}_{\text{Dis}}$.

%%%%%%%%%%%%%%%%%%%%%%%%%%%%%%%%%%%%%%%%%%%%%%%%%%%%%%%%%%%%%%%%%%%%%%%%%%%%%%%%%%%%%%%%%%%%%%%%%%
\subsection{Hard Sample-Conditioned Disentanglement}  \label{sec:hard-dis}

%While existing attention disentanglement methods~\cite{DBLP:conf/cvpr/BhattacharyyaCN23} apply uniform disentanglement across all training samples, hard samples are known to contribute disproportionately to FR optimization and to require dedicated mining or weighting strategies~\cite{DBLP:conf/cvpr/HuangWT0SLLH20, DBLP:conf/cvpr/SchroffKP15, DBLP:conf/cvpr/Kim0L22}. Motivated by this, we condition attention head disentanglement on sample difficulty, encouraging head specialization on hard samples, as demonstrated in Section~\ref{sec:res_approach2}.

While existing attention disentanglement methods~\cite{DBLP:conf/cvpr/BhattacharyyaCN23} apply uniform disentanglement across all training samples, hard samples are known to contribute disproportionately to FR optimization and to require dedicated mining or weighting strategies~\cite{DBLP:conf/cvpr/HuangWT0SLLH20, DBLP:conf/cvpr/SchroffKP15, DBLP:conf/cvpr/Kim0L22}. Motivated by this, we condition attention head disentanglement on sample difficulty, encouraging head specialization on hard samples (Section~\ref{sec:res_approach2}).

Let $e_b$ denote the normalized embedding of sample $b$, $y_b$ its true class label, $w_c$ the normalized center of class $c$, and $s_{b,c} = \langle e_b, w_c \rangle$ the cosine similarity of sample $b$ to center $c$. We categorize samples by whether the true class center is the closest. Wrong samples are misclassified i.e., $\max_{c \neq y_b} s_{b,c} > s_{b,y_b}$. Among the correctly classified samples, easy samples lie far from any negative center, whereas hard samples have high similarity to the closest negative center, creating embedding ambiguity that requires stronger specialization to resolve.

We propose to condition attention head disentanglement on sample hardness, assigning stronger disentanglement to hard samples according to their similarity to the closest negative identity (i.e., the wrong identity with the highest similarity). For a sample \rev{$b$}, we define its hardness as the maximum cosine similarity to a negative class center:

\vspace{-2mm}
\begin{equation}
\rev{h_b = \max_{c \neq y_b} s_{b,c}.}
\end{equation}
\vspace{-3mm}

\rev{The maximum is computed over all negative class centers $\mathbf{w}_c$. Wrong samples are excluded from the weighting to avoid amplifying potential label noise or annotation errors in the training data. We therefore map the hardness to $[0,1]$ and set the weight of wrong samples to zero:}
\begin{equation}
\rev{\tilde{h}_b =
\begin{cases}
\dfrac{h_b + 1}{2}, & \text{if } s_{b,y_b} > \max_{c \neq y_b} s_{b,c} \\[6pt]
0, & \text{otherwise (wrong samples)}
\end{cases}}
\end{equation}
\rev{The per-sample weight is obtained by normalizing $\tilde{h}_b$ by its batch mean, and weights the per-sample disentanglement penalty $\mathcal{L}_{\text{Dis}}^{(b)}$ (Section~\ref{sec:head-disentang}):}

\vspace{-2mm}
\begin{equation}
\rev{w_b = \frac{\tilde{h}_b}{\frac{1}{B}\sum_{b'=1}^{B} \tilde{h}_{b'} + \epsilon}, \qquad
\mathcal{L}_{\text{HarDis}} = \frac{1}{B} \sum_{b=1}^B w_b \cdot \mathcal{L}_{\text{Dis}}^{(b)},}
\end{equation}
\vspace{-3mm}

%\rev{The weight $w_b$ is computed from the current embeddings and class centers but treated as a constant with respect to the decorrelation loss, so no gradient is propagated through it and the feature representation is reshaped only by $\mathcal{L}_{\text{FR}}$. Since $w_b$ is normalized to unit mean over the batch, ViT-HarDis and the uniform variant (ViT-Dis) share the same average penalty magnitude at a given $\lambda$, with the weighting redistributing this magnitude toward harder samples.}
\rev{where $\epsilon$ is a small constant for numerical stability. The weight $w_b$ is computed from the current embeddings and class centers, but is detached and treated as a constant. No gradient is propagated through $w_b$ into $\mathbf{e}_b$ or $\mathbf{w}_c$, so the hardness estimate acts as a per-sample scaling factor. Gradients from $\mathcal{L}_{\text{HarDis}}$ reach the backbone only through the attention maps.}

%\rev{Since $w_b$ is normalized to unit mean over the batch, ViT-HarDis and the uniform variant (ViT-Dis) share the same average penalty magnitude at a given $\lambda$, with the weighting redistributing this magnitude toward harder samples.}

%\rev{Since $w_b$ has unit mean over the batch, the weighting adds no rescaling and redistributes the same average penalty toward harder samples. As the weights are larger where the per-sample penalty is larger, the realized penalties of ViT-HarDis and ViT-Dis are not strictly identical.}
Since $w_b$ has (approximately) unit mean over the batch, the weighting does not rescale the overall penalty but redistributes it toward harder samples. Because the weights may correlate with the per-sample penalty, the realized penalty of ViT-HarDis can still differ from that of ViT-Dis.

%This formulation explicitly emphasizes hard samples by enforcing stronger attention head disentanglement in regions of the feature space where inter-class confusion is highest, thereby prioritizing samples that lie closer to decision boundaries and are more likely to induce misclassification.
This loss is combined with a margin-penalty-based objective \cite{DBLP:conf/cvpr/Kim0L22, DBLP:conf/cvpr/DengGXZ19, DBLP:conf/cvpr/BoutrosDKK22, wang2018cosfacelargemargincosine} to form the total training loss:

\vspace{-2mm}
\begin{equation}
\mathcal{L}_{\text{total}} = \mathcal{L}_{\text{FR}} + \lambda \mathcal{L}_{\text{HarDis}},
\end{equation}
\vspace{-4mm}

where $\lambda$ is the disentanglement weight factor and $\mathcal{L}_{\text{FR}}$ is the margin-penalty based loss. This formulation encourages each attention head to specialize in distinct patterns by \rev{penalizing overlap between the attention maps of different heads}, while the FR loss drives the model to learn discriminative identity embeddings \cite{DBLP:conf/cvpr/Kim0L22, DBLP:conf/cvpr/DengGXZ19, DBLP:conf/cvpr/BoutrosDKK22, wang2018cosfacelargemargincosine}.
\rev{The disentanglement penalty can be applied at any of the $L$ transformer blocks. In our case, $\mathcal{L}_{\text{HarDis}}$ is summed over the selected blocks. An ablation on which blocks to apply the disentanglement to is presented in Section~\ref{sec:selective_dis}.}

%% file: sec/4_experimnetal_setup.tex
%%%%%%%%%%%%%%%%%%%%%%%%%%%%%%%%%%%%%%%%%%%%%%%%%%%%%%%%%%%%%%%%%%%%%%%%%%%%%%%%%%%%%%%%%%%%%%%%%%
\vspace{-1mm}
\section{Experimental Setup} \label{sec:setup}
\vspace{-1mm}
%%%%%%%%%%%%%%%%%%%%%%%%%%%%%%%%%%%%%%%%%%%%%%%%%%%%%%%%%%%%%%%%%%%%%%%%%%%%%%%%%%%%%%%%%%%%%%%%%%
\textbf{Training Datasets:} 
%To train our ViT models,  we employ the MS1MV2 \cite{DBLP:conf/cvpr/DengGXZ19}, MS1MV3 \cite{DBLP:conf/iccvw/DengGZDLS19}, and  WebFace4M \cite{zhu2021webface260mbenchmarkunveilingpower} datasets.
%To train our ViT models,  we employ the MS1MV3 \cite{DBLP:conf/iccvw/DengGZDLS19} and WebFace4M \cite{zhu2021webface260mbenchmarkunveilingpower} datasets. These datasets are widely used in FR research, offering a large and diverse set of identities and images. We use them to train our model to enable a wider comparison to SOTA approaches. MS1MV3 is a cleaned and updated iteration of MS-Celeb-1M \cite{guo2016ms}, containing 5.2M images of 96k identities. WebFace4M is a subset of  WebFace260M \cite{zhu2021webface260mbenchmarkunveilingpower}, consisting of 200K identities and 4 million images. All datasets are provided with aligned and resized images (112×112), obtained using five facial landmarks predicted by RetinaFace \cite{DBLP:conf/cvpr/DengGVKZ20}, following \cite{DBLP:conf/cvpr/KimS0JL24}. For ablation studies, we use a subset of MS1MV3 containing 10K randomly selected identities and approximately 600K images, denoted MS1MV3$_{10\mathrm{K}}$.
We train our ViT models on MS1MV3 \cite{DBLP:conf/iccvw/DengGZDLS19} and WebFace4M \cite{zhu2021webface260mbenchmarkunveilingpower}, two datasets widely used in FR research. MS1MV3 is a cleaned iteration of MS-Celeb-1M \cite{guo2016ms}, containing 5.2M images of 93K identities, while WebFace4M is a subset of WebFace260M \cite{zhu2021webface260mbenchmarkunveilingpower} with 4M images of 200K identities. All images are aligned and resized to 112$\times$112 using five facial landmarks predicted by RetinaFace \cite{DBLP:conf/cvpr/DengGVKZ20}, following \cite{DBLP:conf/cvpr/KimS0JL24}. For ablation studies, we use a subset of MS1MV3 containing 10K randomly selected identities and approximately 600K images, denoted MS1MV3$_{10\mathrm{K}}$. \rev{This subset keeps the ablation studies computationally tractable, while the SOTA comparison in Section~\ref{sec:sota} uses the full datasets.}

%%%%%%%%%%%%%%%%%%%%%%%%%%%%%%%%%%%%%%%%%%%%%%%%%%%%%%%%%%%%%%%%%%%%%%%%%%%%%%%%%%%%%%%%%%%%%%%%%%
\textbf{Evaluation Datasets:} 
%We evaluate the performance of our models on several FR benchmarks. These include LFW \cite{huang:inria-00321923}, CFP-FP \cite{c3517bca662f4193a58fd8f9e3145c8f}, AgeDB30 \cite{moschoglou2017agedb}, CA-LFW \cite{DBLP:journals/corr/abs-1708-08197}, and CP-LFW \cite{CPLFWTech}. We report verification accuracies (\%) following the official evaluation protocols for each of these benchmarks. In addition, we evaluated on large-scale evaluation benchmarks, IJB-B \cite{inproceedingsijbb}, and IJB-C \cite{DBLP:conf/icb/MazeADKMO0NACG18}. For IJB-C and IJB-B, we used the official 1:1 mixed verification protocol and reported the verification performance as true acceptance rates (TAR) at false acceptance rates (FAR) of $1e-4$ and $1e-5$. These benchmarks were selected because they are commonly used to evaluate the latest advancements in FR and offer a diverse range of use cases \cite{Deng_2022, wang2018cosfacelargemargincosine, DBLP:conf/cvpr/BoutrosDKK22, DBLP:conf/iccv/DanLXD0XS23}. We also evaluate our model on the more challenging TinyFace \cite{DBLP:conf/accv/ChengZG18} benchmark, which consists of unconstrained, low-resolution face images. Through this evaluation, we assess the model’s robustness on a low-quality face dataset, highlighting its ability to generalize beyond controlled scenarios.
We evaluate our models on LFW \cite{huang:inria-00321923}, CFP-FP \cite{c3517bca662f4193a58fd8f9e3145c8f}, AgeDB30 \cite{moschoglou2017agedb}, CA-LFW \cite{DBLP:journals/corr/abs-1708-08197}, and CP-LFW \cite{CPLFWTech}, reporting verification accuracies (\%) following the official protocol of each benchmark. We further evaluate on the large-scale IJB-B \cite{inproceedingsijbb} and IJB-C \cite{DBLP:conf/icb/MazeADKMO0NACG18} benchmarks using the official 1:1 mixed verification protocol, reporting true acceptance rates (TAR) at false acceptance rates (FAR) of $10^{-4}$ and $10^{-5}$. These benchmarks are commonly used to evaluate recent advances in FR and cover a diverse range of use cases \cite{Deng_2022, wang2018cosfacelargemargincosine, DBLP:conf/cvpr/BoutrosDKK22, DBLP:conf/iccv/DanLXD0XS23}. Finally, we evaluate on TinyFace \cite{DBLP:conf/accv/ChengZG18}, which consists of unconstrained low-resolution face images, to assess robustness beyond controlled scenarios.

%%%%%%%%%%%%%%%%%%%%%%%%%%%%%%%%%%%%%%%%%%%%%%%%%%%%%%%%%%%%%%%%%%%%%%%%%%%%%%%%%%%%%%%%%%%%%%%%%%
\textbf{Model Architecture:} %We employ two ViT backbones: ViT-S (Small) and ViT-B (Base), with all models using an input image size of $112 \times 112$ and patch size of 8. ViT-S is used for ablation studies and has 12 transformer layers with 8 attention heads, though we explore variants with different numbers of attention heads to study their effect on disentanglement. ViT-B, configured with 24 transformer layers and 16 attention heads, serves as the primary architecture for comparison with state-of-the-art methods in Section \ref{}.
As our base model, we employ ViT with KeyPoint Relative Position Encoding (KP-RPE) \cite{DBLP:conf/cvpr/KimS0JL24}, which leverages keypoints (e.g., facial landmarks) to make ViT more resilient to scale, translation, and pose variations, \rev{making it a strong ViT-based FR baseline}.
%, improving the recognition performance across unconstrained face datasets such as TinyFace \cite{DBLP:conf/accv/ChengZG18}. 
We explore two ViT-KP-RPE backbones: ViT-S (Small) and ViT-B (Base), with all models using an input image size of $112 \times 112$ and a patch size of 8. ViT-S is used for ablation studies and has 12 transformer layers. \rev{We study the effect of different numbers of attention heads (4, 8, 16, and 32) in Section~\ref{sec:res_heads1}, and adopt 16 heads for the remaining ablations.} ViT-B, configured with 24 transformer layers and 16 attention heads, serves as the primary architecture.

%%%%%%%%%%%%%%%%%%%%%%%%%%%%%%%%%%%%%%%%%%%%%%%%%%%%%%%%%%%%%%%%%%%%%%%%%%%%%%%%%%%%%%%%%%%%%%%%%%
\textbf{Training Settings:}
Following \cite{DBLP:conf/cvpr/KimS0JL24}, we employ the AdaFace loss function \cite{DBLP:conf/cvpr/Kim0L22}. Optimization is carried out using AdamW  \cite{loshchilov2019decoupledweightdecayregularization} with a weight decay of $0.05$. The model is trained for $40$ epochs using a batch size of $1024$. We set the initial learning rate to $0.001$ and adopt a cosine learning rate schedule \cite{DBLP:conf/iclr/LoshchilovH17} with a warmup period of $3$ epochs.  We apply a set of data augmentation techniques, following \cite{DBLP:conf/cvpr/KimS0JL24}.
%, including horizontal flipping, brightness, contrast adjustments, scaling, translation, RandAugment \cite{Randaugment_CVPR}, Gaussian blur, cutout, and rotations up to $20^\circ$. Following prior work \cite{Deng_2022, DBLP:conf/cvpr/BoutrosDKK22, DBLP:conf/cvpr/DengGXZ19}, 
%we monitor model convergence after each epoch using several face verification benchmarks, including LFW \cite{huang:inria-00321923}, CALFW \cite{DBLP:journals/corr/abs-1708-08197}, CPLFW \cite{CPLFWTech}, CFP-FP \cite{c3517bca662f4193a58fd8f9e3145c8f}, and AgeDB-30 \cite{moschoglou2017agedb}.

%% file: sec/5_results.tex
\vspace{-1mm}
\section{Results} \label{sec:results}
\vspace{-1mm}
%In this section, we validate our proposed approach, ViT-HarDis, which conditions attention head disentanglement on sample difficulty. First, we investigate the redundancy in multi-head attention by analyzing how the number of heads affects FR performance and inter-head similarity (Section \ref{sec:res_heads1}), finding that additional heads introduce redundancy without improving performance. We then demonstrate the benefits of hard sample-conditioned disentanglement in Section \ref{sec:res_approach2}, showing substantial improvements over both the baseline and uniform disentanglement. In Section \ref{sec:selective_dis}, we study the optimal layer selection strategy, finding that selective disentanglement of the final 25\% of layers outperforms full-network application by a significant margin. Finally, we compare ViT-HarDis against state-of-the-art methods in Section \ref{sec:sota} to demonstrate its competitiveness with existing approaches.

In this section, we validate ViT-HarDis, which conditions attention head disentanglement on sample difficulty. We first investigate redundancy in multi-head attention by analyzing how the number of heads affects FR performance and inter-head similarity (Section~\ref{sec:res_heads1}), finding that additional heads introduce redundancy without improving performance. Section~\ref{sec:res_approach2} then demonstrates the benefits of hard sample-conditioned disentanglement over both the baseline and uniform disentanglement. In Section~\ref{sec:selective_dis}, we study the layer selection strategy, finding that disentangling only the final layers outperforms full-network application. Section~\ref{sec:weight_ablation} ablates the disentanglement weight $\lambda$, and Section~\ref{sec:sota} compares ViT-HarDis against SOTA methods.

% Add computational cost of each settingsfb31puvo Boutros27 July, 12:07 pmis it the case for relatively larger architectures?fb31puvo Boutros27 July, 12:11 pm can we add plots of compuational vs FR performance vs head correlate
%%%%%%%%%%%%%%%%%%%%%%%%%%%%%%%%%%%%%%%%%%%%%%%%%%%%%%%%%%%%%%%%%%%%%%%%%%%%%%%%%%%%%%%%%%%%%%%%%%
\begin{table}[!t]
\centering
\resizebox{1\linewidth}{!}{%
\begin{tabular}{c|cccccc|c|c}
\multirow{2}{*}{\centering \#AH} 
& \multirow{2}{*}{\centering LFW} 
& \multirow{2}{*}{\centering CFP-FP}
& \multirow{2}{*}{\centering AgeDB30} 
& \multirow{2}{*}{\centering CALFW}
& \multirow{2}{*}{\centering CPLFW} 
& \multirow{2}{*}{\centering Avg.}
& \multirow{2}{*}{\centering IJB-B} 
& \multirow{2}{*}{\centering IJB-C} \\
%& \multicolumn{2}{|c}{\rule{0pt}{3.0ex}TinyFace} \\ 
\hhline{~~~~~~~|~|~}
 & & & & & & & &  \\ \midrule
4 & 99.42 & 95.11 & 94.17 & 94.28 & 90.48 & 94.69 & 84.54 & 88.54 \\ 
8 & 99.50 & 95.74 & 94.43 & 94.25 & 90.77 & 94.94 & 84.47 & 88.14 \\	
16 & 99.37 & 95.54 & 94.72 & 94.47 & 90.55 & 94.93 & 84.46 & 88.18 \\ 
32 & 99.58 & 95.00 & 94.62 & 94.17 & 90.85 & 94.84 & 84.19 & 88.25 \\ 
\end{tabular}}
\vspace{-1mm}
\caption{Verification performance ($\%$) of ViT-KP-RPE Small backbones trained on MS1MV3$_{10\mathrm{K}}$ with varying numbers of attention heads (4, 8, 16, and 32) evaluated on several evaluation benchmarks defined in Section \ref{sec:setup}. On IJB-B and IJB-C, results are reported as TAR at FAR of $10^{-4}$. \rev{The verification performance is largely stable across configurations, with no consistent gain from additional attention heads, indicating that the number of heads has limited impact on verification performance.} 
%The consistent performance across configurations demonstrates that the number of attention heads has minimal impact on verification performance.
}
\vspace{-4mm}
\label{tab:heads}
\end{table}

%%%%%%%%%%%%%%%%%%%%%%%%%%%%%%%%%%%%%%%%%%%%%%%%%%%%%%%%%%%%%%%%%%%%%%%%%%%%%%%%%%%%%%%%%%%%%%%%%%
\begin{figure}[!tp]
  \centering
   \includegraphics[width=0.9\linewidth]{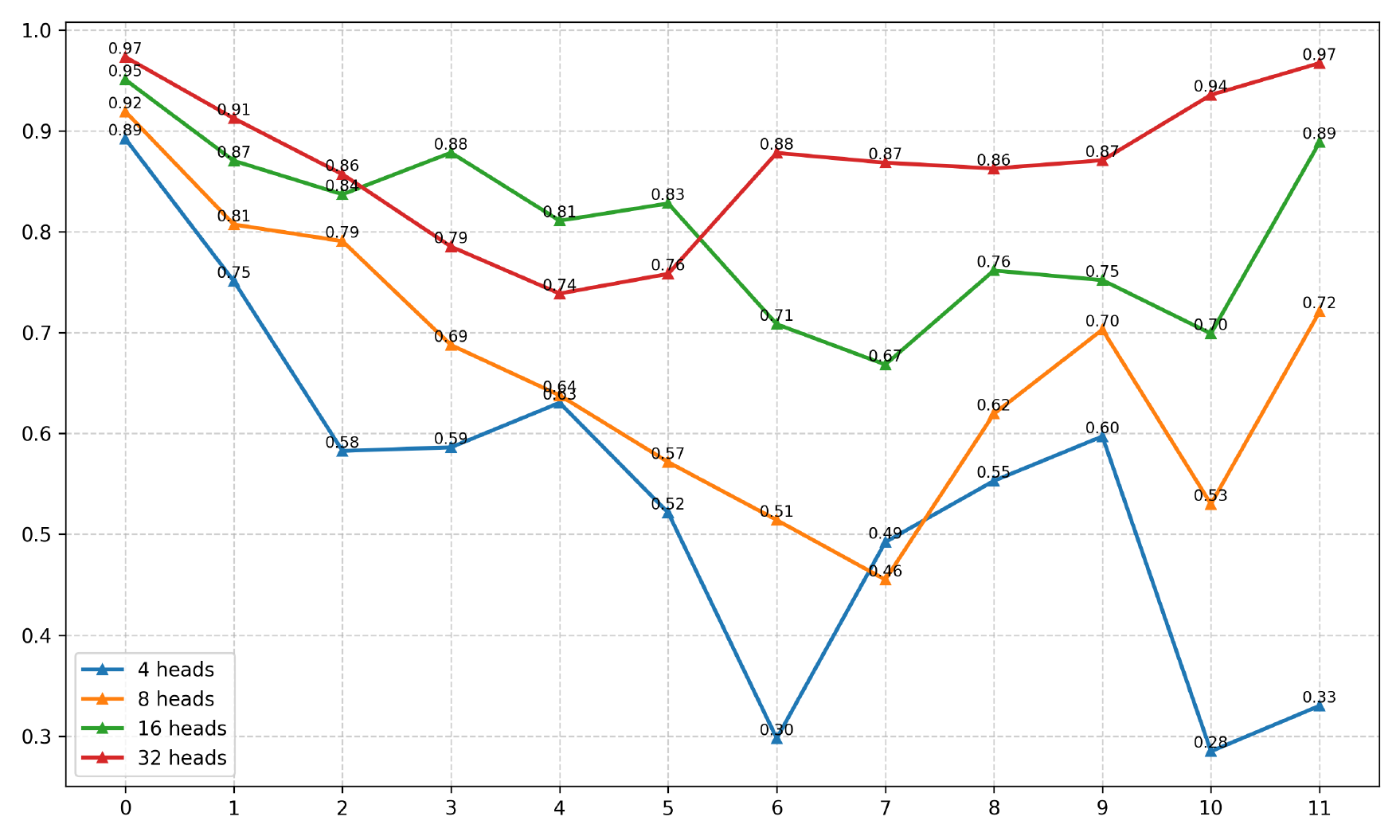}
   %\caption{inter-head attention cosine similarity across transformer depths for ViT-KP-RPE-Small (12 layers). The x-axis shows layer depth, and the y-axis reports the maximum cosine similarity between attention heads within each layer. Results are shown for models with different numbers of attention heads (4, 8, 16, and 32), corresponding to blue, orange, green, and red curves, respectively. We observe that models with more attention heads generally exhibit higher inter-head similarity across most depths, indicating increased head redundancy as the number of heads increases.}
   \vspace{-4mm}
   \caption{Inter-head attention cosine similarity across transformer depths for ViT-KP-RPE Small (12 layers). The y-axis reports the maximum cosine similarity between attention heads within each layer, for models with 4, 8, 16, and 32 heads (blue, orange, green, and red, respectively). Models with more attention heads generally exhibit higher similarity across most depths, indicating increased head redundancy as the number of heads increases.}
   \label{fig:Attention_heads_redundancy}
   \vspace{-4mm}
\end{figure}

%%%%%%%%%%%%%%%%%%%%%%%%%%%%%%%%%%%%%%%%%%%%%%%%%%%%%%%%%%%%%%%%%%%%%%%%%%%%%%%%%%%%%%%%%%%%%%%%%%
\subsection{Attention Head Redundancy in ViT-based FR} \label{sec:res_heads1}
In this section, we investigate how the number of attention heads affects FR performance in ViT-KP-RPE. Accordingly, we evaluate four configurations (4, 8, 16, and 32 heads), with verification accuracy results presented in Table \ref{tab:heads}, and analyze the resulting attention patterns across depths in Figure \ref{fig:Attention_heads_redundancy}. We make the following observations:

\begin{itemize}
  %\item From Table \ref{tab:heads}, which reports verification accuracy across the considered configurations on several evaluation benchmarks defined in Section \ref{sec:setup}, we observe that the model with the fewest attention heads achieves the best performance on both large-scale benchmarks, namely IJB-B and IJB-C\rev{, although the differences across configurations are small and the ranking varies on other benchmarks. Overall, additional attention heads do not yield consistent improvements in verification performance.}
  \item From Table~\ref{tab:heads}, which reports verification accuracy across the considered configurations on several evaluation benchmarks defined in Section~\ref{sec:setup}, we observe that verification performance is largely stable across configurations. \rev{The model with the fewest attention heads achieves the best results on both large-scale benchmarks, IJB-B and IJB-C, but the differences are small and the ranking varies on other benchmarks. Overall, additional attention heads yield no consistent gain, indicating that the number of heads has limited impact on verification performance.}
  \item From Figure \ref{fig:Attention_heads_redundancy}, which presents the maximum inter-head attention cosine similarity across transformer depths for the considered configurations, we observe that as the number of attention heads increases, inter-head cosine similarity rises across most depths. This suggests that additional heads do not develop specialized roles, but instead learn redundant attention patterns.
\end{itemize}

\rev{Together, the minimal performance variance in Table~\ref{tab:heads} and the high inter-head redundancy in Figure~\ref{fig:Attention_heads_redundancy} indicate that additional heads are not effectively used. This motivates head disentanglement to enforce meaningful specialization.} 
\rev{For all remaining ablations in Sections~\ref{sec:res_approach2}--\ref{sec:weight_ablation}, we adopt ViT-S with 16 attention heads, matching the head count of the ViT-B backbone used in our SOTA comparison and exhibiting clear inter-head redundancy.}

%The minimal performance variance across different head configurations in Table \ref{tab:heads}, combined with high inter-head redundancy at different depths observed in Figure \ref{fig:Attention_heads_redundancy}, demonstrates that the current configurations do not effectively leverage attention head diversity. Also, the model with the fewest heads and the lowest inter-head similarity achieves the best results on large-scale benchmarks, indicating that it does not benefit from redundancy. This motivates our approach of exploring head disentanglement to enforce meaningful specialization among attention heads. \rev{For all subsequent experiments, we adopt ViT-S with 16 attention heads, matching the ViT-B backbone used in our SOTA comparison and exhibiting clear inter-head redundancy.}

%%%%%%%%%%%%%%%%%%%%%%%%%%%%%%%%%%%%%%%%%%%%%%%%%%%%%%%%%%%%%%%%%%%%%%%%%%%%%%%%%%%%%%%%%%%%%%%%%%
\begin{table*}[!t]
\centering
\resizebox{0.75\linewidth}{!}{%
\begin{tabular}{c|cccccc|cc|cc|cc}
\multirow{2}{*}{\centering Method} 
& \multirow{2}{*}{\centering LFW} 
& \multirow{2}{*}{\centering CFP-FP} 
& \multirow{2}{*}{\centering AgeDB30} 
& \multirow{2}{*}{\centering CALFW} 
& \multirow{2}{*}{\centering CPLFW} 
& \multirow{2}{*}{\centering Avg.} 
& \multicolumn{2}{|c|}{\rule{0pt}{3.0ex}IJB-B}
& \multicolumn{2}{|c|}{\rule{0pt}{3.0ex}IJB-C} 
& \multicolumn{2}{|c}{\rule{0pt}{3.0ex}TinyFace} \\ 
\hhline{~~~~~~~|--|--|--}
 & & & & & & & $10^{-4}$ & $10^{-5}$ & $10^{-4}$ & $10^{-5}$ &  Rank-1 & Rank-5 \\ \midrule
ViT & 99.37 & 95.54 & 94.72 & 94.47 & 90.55 & 94.93 & 84.46 & 72.16 & 88.18 & 78.25 & 66.12 & 71.27  \\ %	
ViT-Dis & 99.43 & 95.61 & 94.48 & 94.30 & 90.72 & 94.91 & 85.03 & 69.82 & 88.28 & 78.51 & 66.74 & 71.70 \\ %	
%ViT-HarDis & 99.36	&95.57	&95.10	&94.40	&90.85	&\textbf{95.06}	&\textbf{85.72}	&\textbf{74.29}	&\textbf{89.09}	&\textbf{79.93}	&\textbf{67.30}	&\textbf{72.29} \\ %
\rev{ViT-HarDis$^\dagger$} & 99.43	& 95.36	& 94.82	& 94.52	&90.72	&94.97	&85.07	&\textbf{73.17}	&88.59	&79.51	&\textbf{67.36}	&71.89 \\ %
ViT-HarDis & 99.47 & 95.67	& 94.90	& 94.47	& 90.67	& \textbf{95.04}	&\textbf{85.72}	&71.74	&\textbf{89.09}	&\textbf{80.52}	&67.11	&\textbf{72.13} \\ %	
\end{tabular}}
\vspace{-2mm}
\caption{Verification performance (\%) of ViT-KP-RPE Small trained on MS1MV3$_{10\mathrm{K}}$ comparing the baseline model with two disentanglement variants: uniform disentanglement (ViT-Dis) and our proposed hard sample-conditioned disentanglement (ViT-HarDis). On IJB-B and IJB-C, results are reported as TAR at FAR of $10^{-4}$ and $10^{-5}$. \rev{ViT-HarDis$^\dagger$ denotes a variant of ViT-HarDis in which misclassified (wrong) samples are not excluded from the disentanglement weighting.} Our approach, ViT-HarDis, outperforms both ViT and ViT-Dis on most considered benchmarks, demonstrating the effectiveness of sample-aware regularization.}
\label{tab:decor}
\vspace{-3mm}
\end{table*}

%%%%%%%%%%%%%%%%%%%%%%%%%%%%%%%%%%%%%%%%%%%%%%%%%%%%%%%%%%%%%%%%%%%%%%%%%%%%%%%%%%%%%%%%%%%%%%%%%%
\begin{figure*}[!tp]
    \centering

    \begin{subfigure}[t]{0.33\linewidth}
        \centering
        \includegraphics[width=\linewidth]{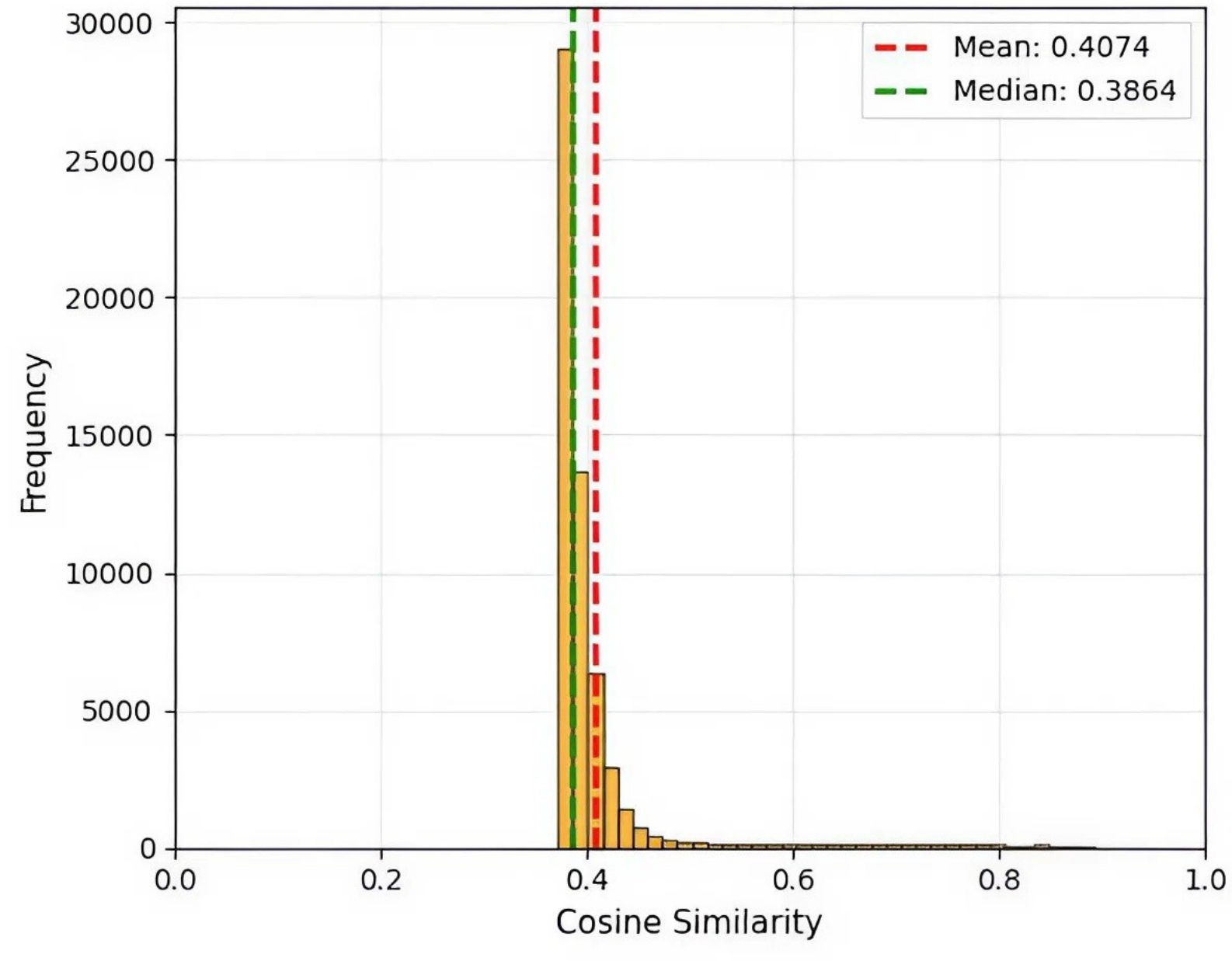}
        \caption{\rev{ViT}}
    \end{subfigure}
    \hfill
    \begin{subfigure}[t]{0.33\linewidth}
        \centering
        \includegraphics[width=\linewidth]{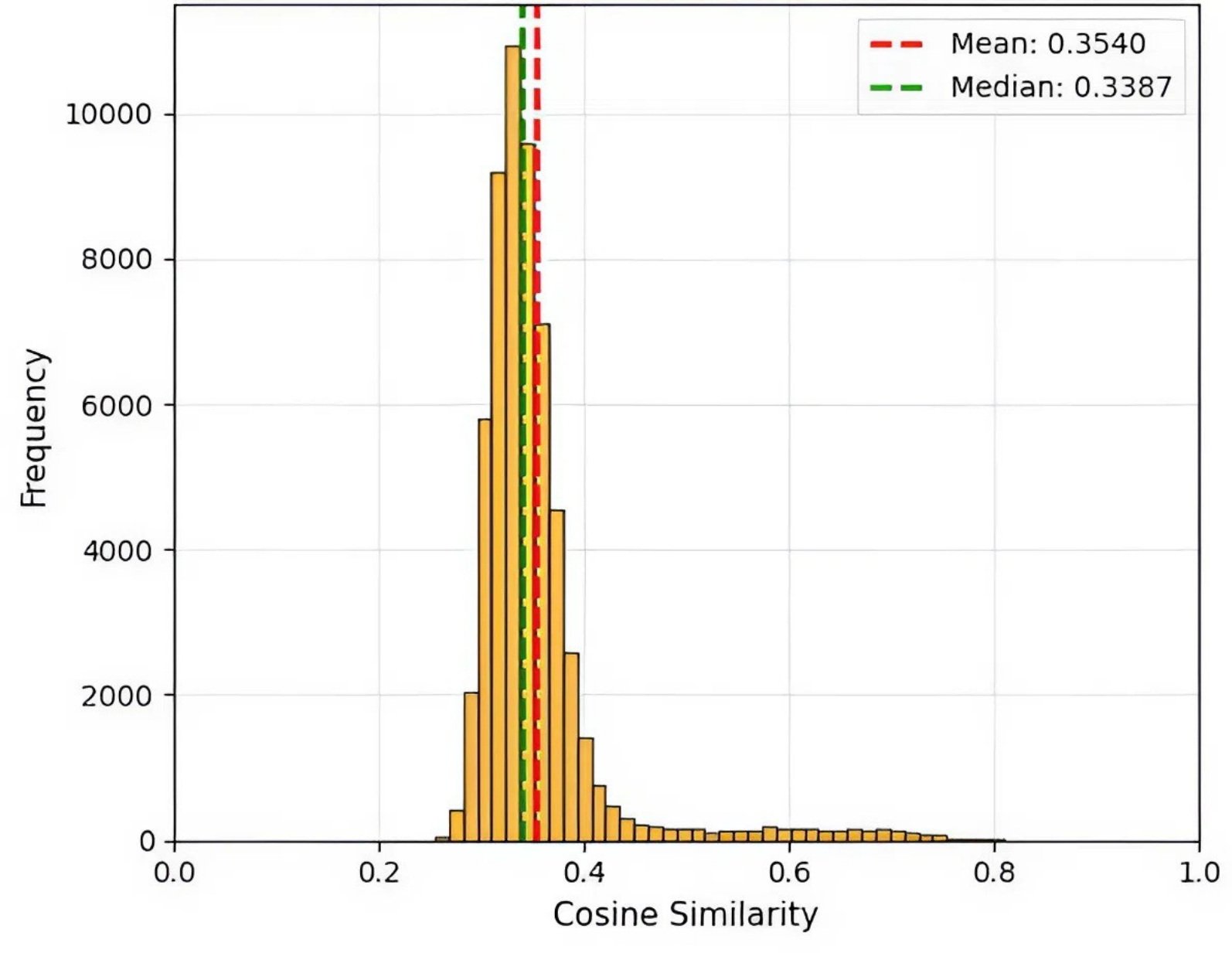}
        \caption{\rev{ViT-Dis}}
    \end{subfigure}
    \hfill
    \begin{subfigure}[t]{0.33\linewidth}
        \centering
        \includegraphics[width=\linewidth]{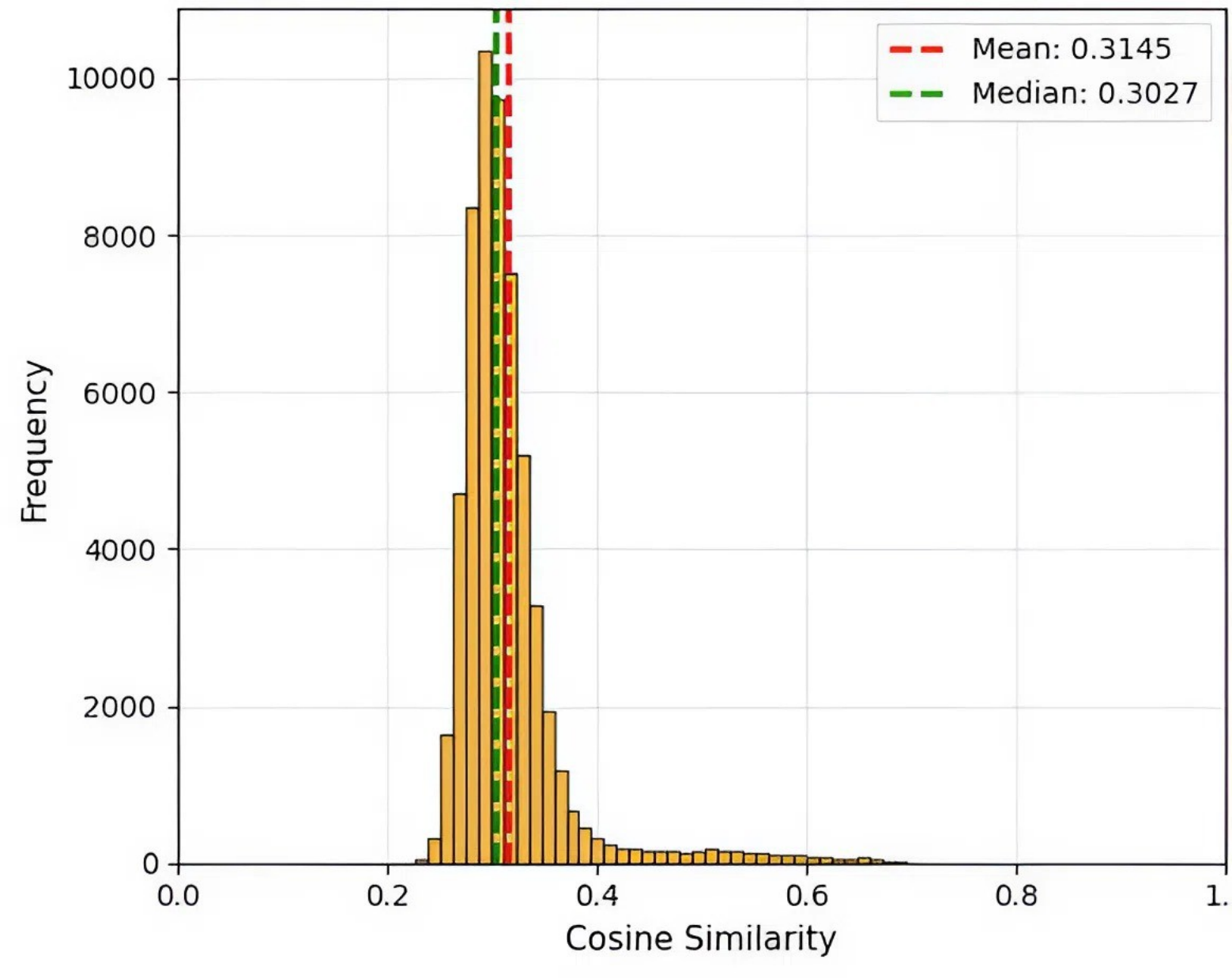}
        \caption{\rev{ViT-HarDis}}
    \end{subfigure}
    \vspace{-3mm}
    \caption{Distribution of cosine similarity to the closest negative class center for hard samples (top 10\% by baseline model) from MS1MV3$_{10\mathrm{K}}$, across ViT-KP-RPE, ViT-Dis, and ViT-HarDis. Red and green dashed lines indicate the mean and median. ViT-HarDis achieves the lowest mean similarity (0.31), reducing identity ambiguity by adaptively specializing attention heads.}
    \label{fig:distribution}
    \vspace{-3mm}
\end{figure*}

%%%%%%%%%%%%%%%%%%%%%%%%%%%%%%%%%%%%%%%%%%%%%%%%%%%%%%%%%%%%%%%%%%%%%%%%%%%%%%%%%%%%%%%%%%%%%%%%%%
\begin{figure}[!tp]
  \centering
   \includegraphics[width=0.9\linewidth]{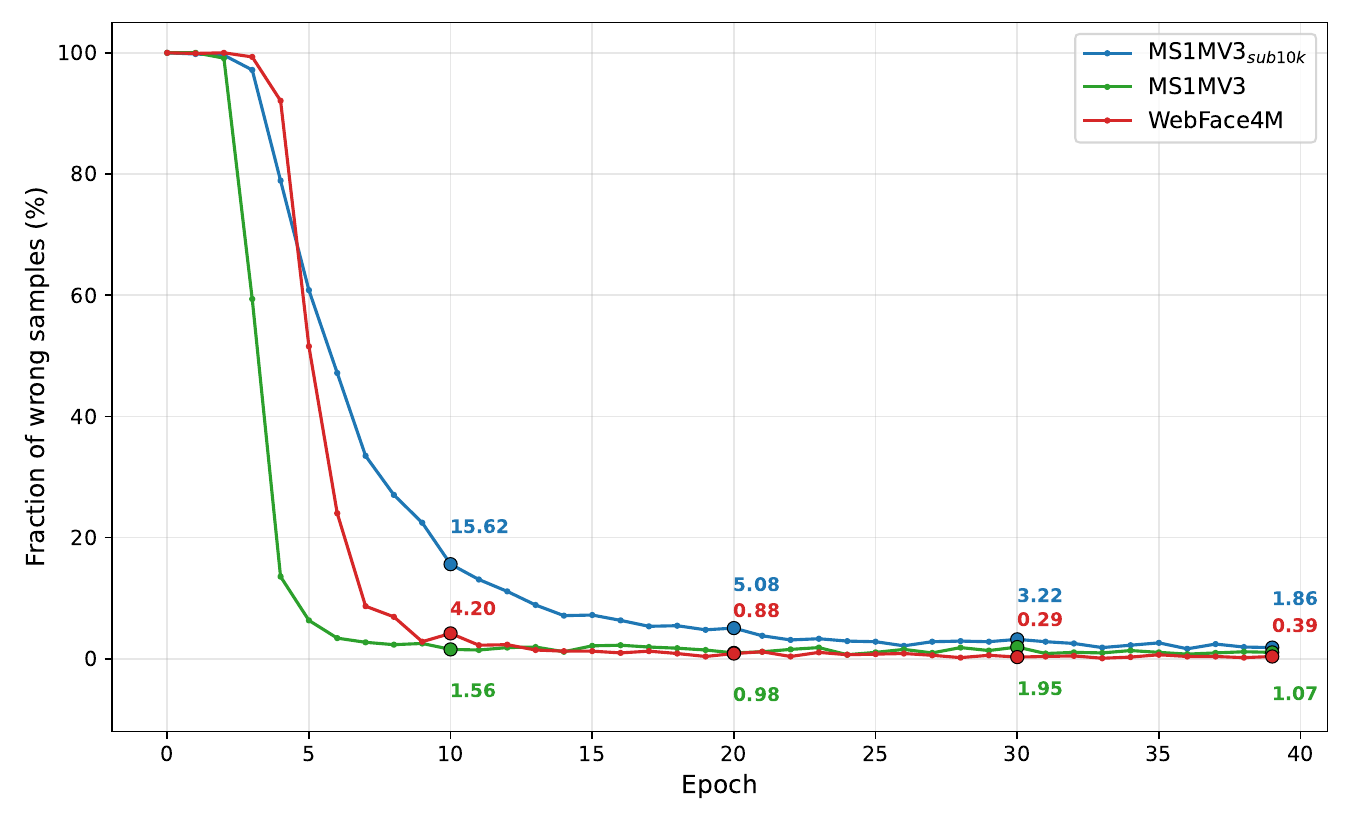}
   \vspace{-3mm}
    \caption{\rev{Fraction of \emph{wrong samples} per training epoch for ViT-HarDis on MS1MV3$_{10\mathrm{K}}$ (0.6M samples), MS1MV3 (5.2M samples), and WebFace4M (4M samples). The y-axis reports the percentage of samples classified as wrong, as defined in Section~\ref{sec:hard-dis}, and therefore excluded from the hardness-weighted disentanglement loss. On MS1MV3$_{10\mathrm{K}}$, samples are reintegrated as classification improves, dropping to about 16\% after ten epochs and below 2\% by the end, while WebFace4M and MS1MV3 converge faster, falling to about 4\% and 2\% after ten epochs.}}
    \vspace{-4mm}
   \label{fig:wrong_samples}
\end{figure}

%%%%%%%%%%%%%%%%%%%%%%%%%%%%%%%%%%%%%%%%%%%%%%%%%%%%%%%%%%%%%%%%%%%%%%%%%%%%%%%%%%%%%%%%%%%%%%%%%%
\begin{figure}[!tp]
    \centering
    \begin{subfigure}[t]{\linewidth}
        \centering
        \includegraphics[width=1\linewidth]{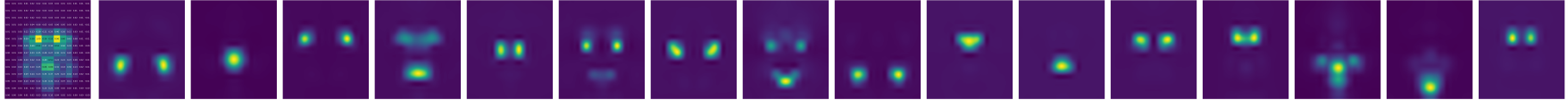}
        \caption{Baseline (\rev{ViT})}
    \end{subfigure}
    \begin{subfigure}[t]{\linewidth}
        \centering
        \includegraphics[width=1\linewidth]{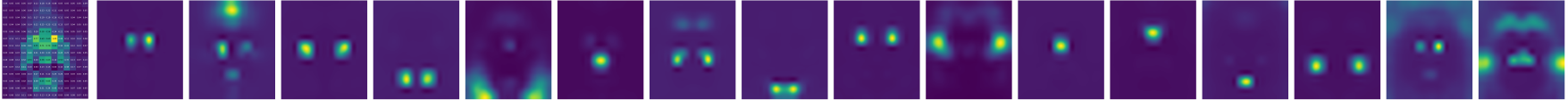}
        \caption{Uniform attention disentanglement (\rev{ViT-Dis})}
    \end{subfigure}
    \begin{subfigure}[t]{\linewidth}
        \centering
        \includegraphics[width=1\linewidth]{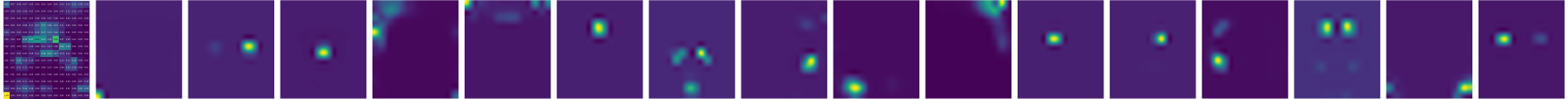}
        \caption{Hard sample-conditioned attention disentanglement (\rev{ViT-HarDis})}
    \end{subfigure}
    \vspace{-2mm}
    \caption{Average attention maps across all hard samples (top 10\% selected by the baseline model) from MS1MV3$_{10\mathrm{K}}$, visualized over 16 attention heads for Baseline ViT-KP-RPE, ViT-Dis, and ViT-HarDis. The first column shows the average over all heads and samples, while the remaining columns show \rev{heads 1–16, ordered left to right}. The comparison highlights differences in specialization and redundancy among the three models.}
    \label{fig:attention}
    \vspace{-6mm}
\end{figure}

% The setup states that ViT-S has eight attention heads, but Figure 3 visualizes 16 heads, and Table 2 does not explicitly state the head count. The exact configurations for Tables 2–3 and Figures 2–3 should be given.

%%%%%%%%%%%%%%%%%%%%%%%%%%%%%%%%%%%%%%%%%%%%%%%%%%%%%%%%%%%%%%%%%%%%%%%%%%%%%%%%%%%%%%%%%%%%%%%%%%
\subsection{Hard Sample-conditioned Disentanglement} \label{sec:res_approach2}
In this section, we validate that the model trained with hard sample-conditioned disentanglement (ViT-HarDis) outperforms both baseline (ViT) and uniform disentanglement (ViT-Dis) on most considered benchmarks. For ViT-Dis and ViT-HarDis, we apply the disentanglement to the final transformer blocks (10--12), which yields the best results as shown in Section~\ref{sec:selective_dis}. We set $\lambda = 1$ for both, and study its effect in Section~\ref{sec:weight_ablation}, which confirms $\lambda = 1$ as the best-performing value on the large-scale benchmarks.

%In this section, we validate that \rev{the} model trained with hard sample-conditioned disentanglement (ViT-HarDis) is superior to both baseline (ViT) and uniform disentanglement (ViT-Dis). We investigate three key aspects: (1) overall face verification performance, (2) the mechanism of hard sample ambiguity reduction, and (3) the underlying attention specialization patterns. By examining verification accuracy, cosine similarity distributions, and attention map visualizations, we provide comprehensive evidence that sample-aware conditioning is essential for effective head specialization. \rev{For ViT-Dis and ViT-HarDis, we apply the disentanglement to the final transformer blocks (10--12), which yields the best results as shown in Section~\ref{sec:selective_dis}. We set the disentanglement weight to $\lambda=1$ for both ViT-Dis and ViT-HarDis, and study its effect in the ablation of Section \ref{sec:weight_ablation}, which confirms $\lambda=1$ as the best-performing value on the large-scale benchmarks.}

\textbf{Face verification performance.}
Table \ref{tab:decor} reports verification accuracy across all three variants on several evaluation benchmarks defined in Section \ref{sec:setup}. We observe that:

\begin{itemize}
\item \rev{At FAR$=10^{-4}$, both disentanglement approaches improve over the baseline, but ViT-Dis provides only modest gains (84.46\% to 85.03\% on IJB-B, 88.18\% to 88.28\% on IJB-C), whereas ViT-HarDis reaches 85.72\% and 89.09\%, corresponding to improvements of 1.26 and 0.91 percentage points. At FAR$=10^{-5}$, ViT-Dis falls below the baseline on IJB-B, while ViT-HarDis and ViT-HarDis$^{\dagger}$ both remain above it.}
\item \rev{ViT-HarDis also outperforms ViT-Dis on the large-scale benchmarks, with gains of 0.69 and 0.81 percentage points on IJB-B and IJB-C at FAR$=10^{-4}$, rising to 1.92 and 2.01 percentage points at FAR$=10^{-5}$, demonstrating the benefit of hard sample-conditioned attention disentanglement for effective head specialization.}
%\item On TinyFace, ViT-HarDis achieves 67.30\% Rank-1 accuracy, significantly outperforming both the baseline (66.12\%) and ViT-Dis (66.74\%), confirming the effectiveness of hard sample-conditioning on challenging low-quality evaluation benchmark.
\item \rev{Early in training, when most samples are still misclassified (Figure~\ref{fig:wrong_samples}), excluding wrong samples concentrates the batch-normalized penalty on the few correctly classified ones, and the penalty vanishes only when no sample in a batch is correctly classified. We therefore evaluate ViT-HarDis$^\dagger$, which does not exclude wrong samples and weights all samples from the first iteration. ViT-HarDis$^\dagger$ still outperforms ViT-Dis on the large-scale benchmarks and TinyFace, indicating that the gain stems from hardness weighting rather than from the exclusion of wrong samples. Compared to ViT-HarDis$^\dagger$, ViT-HarDis achieves higher results on IJB-C at both FARs, on IJB-B at FAR$=10^{-4}$, and on the benchmark average, while ViT-HarDis$^\dagger$ is stronger only on IJB-B at FAR$=10^{-5}$ and on TinyFace Rank-1. This shows that excluding wrong samples further improves performance on most benchmarks, consistent with avoiding the amplification of potential label noise.}
\end{itemize}

\textbf{Influence on hard samples.}
\rev{To assess hard sample ambiguity, we first compute the cosine similarity of all training samples in MS1MV3$_{10\mathrm{K}}$ to their closest negative class center using the baseline ViT-KP-RPE Small, then select the 10\% hardest samples. Figure~\ref{fig:distribution} shows their similarity distribution for all three variants.} We observe that the mean cosine similarity of hard samples to their closest negative class center is 0.40 for the baseline, dropping to 0.35 for uniform disentanglement, and reaching 0.31 for ViT-HarDis. This demonstrates that our approach reduces the similarity of hard samples to competing class centers, thereby lowering embedding ambiguity and improving identity discrimination.
\rev{We further track the fraction of wrong samples, as defined in Section~\ref{sec:hard-dis}, throughout training. Figure~\ref{fig:wrong_samples} shows that it starts near 100\% at initialization, when class centers are untrained, and decreases sharply during training. On MS1MV3$_{10\mathrm{K}}$, about 16\% of samples remain excluded after ten epochs, dropping to roughly 2\% by the end. MS1MV3 converges fastest, falling to about 2\% within ten epochs, while WebFace4M converges more slowly but reaches the lowest final fraction at about 0.4\%.}

\textbf{Attention maps analysis.}
%Figure \ref{fig:attention} visualizes the average attention maps across hard samples (top 10\% from MS1MV3$_{10K}$) for all 16 attention heads in the three variants. Qualitatively, the attention patterns reveal important differences in head specialization. The baseline model and ViT-Dis exhibit substantial redundancy in attention focus, with most heads concentrating on similar facial regions (e.g., both eyes). In contrast, ViT-HarDis demonstrates significantly reduced redundancy, with individual heads specializing on distinct facial regions. This observation is quantified by computing average inter-head attention similarity across all hard samples: the baseline achieves 0.21, ViT-Dis reduces this to 0.09, and ViT-HarDis achieves 0.01. The dramatic reduction in attention similarity for ViT-HarDis indicates that hard sample-conditioned disentanglement successfully enforces meaningful head specialization, where each head learns to focus on complementary facial features rather than redundantly attending to the same regions. This specialized attention distribution directly contributes to improved identity discrimination on hard samples, as demonstrated by the cosine similarity reductions observed in Figure \ref{fig:distribution}.
Figure~\ref{fig:attention} visualizes the average attention maps across hard samples (top 10\% from MS1MV3$_{10K}$) for all 16 attention heads in the three variants. The baseline and ViT-Dis exhibit substantial redundancy, with most heads concentrating on similar facial regions (e.g., both eyes), whereas ViT-HarDis shows reduced redundancy, with individual heads specializing on distinct regions. This is quantified by the average inter-head attention similarity across all hard samples: 0.21 for the baseline, 0.09 for ViT-Dis, and 0.01 for ViT-HarDis. This reduction indicates that hard sample-conditioned disentanglement enforces head specialization, with each head focusing on complementary facial features rather than redundantly attending to the same regions, which contributes to improved identity discrimination on hard samples, as shown in Figure~\ref{fig:distribution}.

\textbf{Computational cost.}
\rev{Our approach does not modify the ViT-KP-RPE backbone and introduces no additional parameters or operations at inference. Inference cost is therefore identical to the baseline, and only training time is affected. All models were trained on 8$\times$ NVIDIA A100 SXM4 40\,GB GPUs, using the parameters defined in Section~\ref{sec:setup}. To quantify the overhead of the additional loss computation, we report the training time of ViT-KP-RPE and ViT-HarDis on WebFace4M with a ViT-B backbone and the disentanglement applied to 6 transformer blocks, over 40 epochs. The baseline required 39 minutes per epoch on average, while ViT-HarDis required 51 minutes, corresponding to a total training time of 26 and 34 hours, respectively, i.e., an increase of ${\sim}30\%$.}

\rev{Together, these analyses show that hard sample-conditioned disentanglement reduces inter-head similarity, which in turn lowers hard sample ambiguity and improves verification, validating sample difficulty as a signal for guiding head specialization.}

%Together, these three analyses provide complementary evidence of why hard sample-conditioned disentanglement outperforms both baseline and uniform disentanglement. Verification improvements on challenging benchmarks (up to 1.26 percentage points on IJB-B) directly correlate with reduced hard sample ambiguity: mean cosine similarity to competing centers drops from 0.40 (baseline) to 0.31 (ViT-HarDis). This ambiguity reduction is enabled by enhanced attention specialization, where inter-head similarity decreases from 0.21 to 0.01. Rather than redundantly attending to the same facial regions, individual heads in ViT-HarDis learn complementary features. These results validate our core hypothesis: sample difficulty is a crucial signal for guiding attention head specialization. By adaptively conditioning regularization on embedding confusability, hard sample-conditioned disentanglement enforces specialization precisely where identity ambiguity is highest, thereby unlocking the latent capacity of transformer-based face recognition models.

%%%%%%%%%%%%%%%%%%%%%%%%%%%%%%%%%%%%%%%%%%%%%%%%%%%%%%%%%%%%%%%%%%%%%%%%%%%%%%%%%%%%%%%%%%%%%%%%%%
\begin{table}[!t]
\centering
\resizebox{1\linewidth}{!}{%
\begin{tabular}{c|cccccc|c|c}
\multirow{2}{*}{\centering Disentangled Layers} 
& \multirow{2}{*}{\centering LFW} 
& \multirow{2}{*}{\centering CFP-FP} 
& \multirow{2}{*}{\centering AgeDB30} 
& \multirow{2}{*}{\centering CALFW} 
& \multirow{2}{*}{\centering CPLFW} 
& \multirow{2}{*}{\centering Avg.} 
& \multirow{2}{*}{\centering IJB-B} 
& \multirow{2}{*}{\centering IJB-C} \\
%& \multicolumn{2}{|c}{\rule{0pt}{3.0ex}TinyFace} \\ 
\hhline{~~~~~~~|~|~}
 & & & & & & & &  \\ \midrule
1--12 (100\%) & 99.42 & 94.79 & 94.18 & 93.78 & 89.63 & 94.36 & 82.20 & 83.94 \\ % & 65.34 & 70.52  \\ %	
\hdashline
4--12 (75\%) & 99.48 & 95.13 & 94.48 & 94.10 & 90.32 & 94.70 & 83.94 & 86.54 \\ % & 66.23 & 71.11 \\ %	
7--12 (50\%) & 99.43 & 95.41 & 94.75 & 94.50 & 90.72 & 94.96 & 84.70 & 87.44 \\ % & 66.74 & 71.78 \\ %
10--12 (25\%) & 99.47	&95.67	&94.90	&94.47	&90.67	&\textbf{95.04}	&\textbf{85.72}	&\textbf{89.09}	\\ % &\textbf{67.30}	&\textbf{72.29} \\ %
\hdashline
1--9 (75\%) & 99.43 & 95.19 & 94.58 & 94.00 & 90.40 & 94.72 & 83.56 & 86.48 \\ % & 65.37 & 70.52 \\ %	
1--6 (50\%) & 99.48 & 95.30 & 94.22 & 94.27 & 90.50 & 94.75 & 84.63 & 88.16 \\ % & 65.56 & 70.55 \\ %
1--3 (25\%) & 99.47 & 95.27 & 94.92 & 94.43 & 90.87 & 94.99 & 85.14 & 88.65 \\ % & 66.87 & 72.08 \\ %
\end{tabular}}
\vspace{-2mm}
\caption{Verification performance ($\%$) of ViT-HarDis (Small backbones) trained on MS1MV3$_{10\mathrm{K}}$ with selective layer disentanglement evaluated on several evaluation benchmarks defined in Section \ref{sec:setup}. On IJB-B and IJB-C, results are reported as TAR at FAR of $10^{-4}$. Results compare two layer selection strategies: disentangling from the end (upper section: 1--12, 4--12, 7--12, 10--12) and consecutive layers from the beginning (lower section: 1--9, 1--6, 1--3), corresponding to 75\%, 50\%, and 25\% of total layers, respectively. The disentanglement of the final 25\% layers (10–12) performs best on average and on the large-scale benchmarks.}
\vspace{-2mm}
\label{tab:layers}
\end{table}

%%%%%%%%%%%%%%%%%%%%%%%%%%%%%%%%%%%%%%%%%%%%%%%%%%%%%%%%%%%%%%%%%%%%%%%%%%%%%%%%%%%%%%%%%%%%%%%%%%
\begin{table}[!t]
\centering
\resizebox{1\linewidth}{!}{%
\begin{tabular}{c|cccccc|c|c}
\multirow{2}{*}{\centering \rev{$\lambda$}} 
& \multirow{2}{*}{\centering \rev{LFW}} 
& \multirow{2}{*}{\centering \rev{CFP-FP}} 
& \multirow{2}{*}{\centering \rev{AgeDB30}} 
& \multirow{2}{*}{\centering \rev{CALFW}} 
& \multirow{2}{*}{\centering \rev{CPLFW}} 
& \multirow{2}{*}{\centering \rev{Avg.}}
& \multirow{2}{*}{\centering \rev{IJB-B}} 
& \multirow{2}{*}{\centering \rev{IJB-C}} \\
%& \multicolumn{2}{|c}{\rule{0pt}{3.0ex}TinyFace} \\ 
\hhline{~~~~~~~|~|~}
 & & & & & & & &  \\ \midrule
\rev{0.5}  & 99.43	&95.70	&94.97	&94.30	&90.97	&\textbf{95.07}  &85.37 &88.97 \\ %
\rev{1}    & 99.47	&95.67	&94.90	&94.47	&90.67	&95.04  &\textbf{85.72} &\textbf{89.09} \\ %
\rev{1.75} & 99.55	&95.66	&94.77	&94.15	&90.58	&94.94  &85.45 &88.79 \\ %
\rev{2.5}  & 99.48	&95.57	&94.35	&94.12	&90.78	&94.86  &85.27 &88.48 \\ %
\rev{5}    & 99.50	&95.20	&94.35	&94.43	&90.23	&94.74  &85.00 &88.31 \\ %
\rev{10}   & 99.40	&95.33	&94.30	&94.25	&90.50	&94.76  &84.33 &87.81 \\ %
\end{tabular}}
\vspace{-2mm}
\caption{\rev{Ablation on the disentanglement weight $\lambda$ for ViT-HarDis. Results on IJB-B and IJB-C are reported as TAR at FAR$=10^{-4}$. $\lambda=1$ achieves the best performance on the large-scale IJB-B and IJB-C benchmarks.}}
%\caption{\rev{Ablation on the disentanglement weight $\lambda$ for ViT-HarDis. Results on IJB-B and IJB-C are reported as TAR (@FAR=$1e-4$). $\lambda=1$ achieves the best performance on the large-scale IJB-B and IJB-C benchmarks.}}
\label{tab:weight_ablation}
\vspace{-4mm}
\end{table}

%%%%%%%%%%%%%%%%%%%%%%%%%%%%%%%%%%%%%%%%%%%%%%%%%%%%%%%%%%%%%%%%%%%%%%%%%%%%%%%%%%%%%%%%%%%%%%%%%%
\begin{table*}[!h]
\centering
\resizebox{0.9\linewidth}{!}{%
  \centering
\begin{tabular}{ccc|cccccc|cc|cc}
\multirow{2}{*}{\centering Method} 
& \multirow{2}{*}{\centering Backbone} 
& \multirow{2}{*}{\centering Train data} 
& \multirow{2}{*}{\centering LFW} 
& \multirow{2}{*}{\centering CFP-FP} 
& \multirow{2}{*}{\centering AgeDB30} 
& \multirow{2}{*}{\centering CALFW} 
& \multirow{2}{*}{\centering CPLFW} 
& \multirow{2}{*}{\centering Avg.} 
& \multirow{2}{*}{\centering IJB-B} 
& \multirow{2}{*}{\centering IJB-C} 
& \multicolumn{2}{c}{\rule{0pt}{3.0ex}TinyFace} 
\\ 
\cline{12-13}
& & & & & & & & & & & Rank-1 & Rank-5 \\ \midrule
 
%CosFace \cite{DBLP:conf/cvpr/WangWZJGZL018} (CVPR 2018) & ResNet101 & MS1MV2 & 99.81 & 98.12 & 98.11 & 95.76 & 92.28 & 96.82 &  94.80 & 96.37 \\  % & - & - \\ 
%ArcFace \cite{DBLP:conf/cvpr/DengGXZ19} (CVPR 2019) & ResNet101 & MS1MV2  & 99.83 & 98.27 & 98.28 & 95.45 & 92.08 & 96.78 & 94.25 &  96.03 \\  % & - & - \\ 
%BroadFace \cite{DBLP:conf/eccv/KimPS20} (ECCV 2020) & ResNet101 & MS1MV2  & \textbf{99.85} & 98.63 & \textbf{98.38} & \textbf{96.20} & 93.17 & \textbf{97.25} & 94.97 & 96.38 \\  % & - & - \\ 
%CurricularFace \cite{DBLP:conf/cvpr/HuangWT0SLLH20} (CVPR 2020) & ResNet101 & MS1MV2  & 99.80 & 98.37 & 98.32 & \textbf{96.20}  & 93.13 & 97.16 & 94.80 &  96.10 \\  % & 63.68 & 67.65 \\ %%adaface keypoint
%URL \cite{DBLP:conf/cvpr/Shi0SC020} (CVPR 2020) & ResNet101 & MS1MV2  & 99.78 & 98.64 & - & - & - & - & - & 96.60 \\  % & 63.89 & 68.67 \\ 
%MagFace \cite{DBLP:conf/cvpr/MengZH021} (CVPR 2021) & ResNet101 & MS1MV2  & 99.83 & 98.46 & 98.17 & 96.15 & 92.87 &  97.10 & 94.51 & 95.97 \\  % & - & - \\ 
%AdaFace \cite{DBLP:conf/cvpr/Kim0L22} (CVPR 2022) & ResNet101 & MS1MV2  & 99.82 & 98.49 & 98.05 & 96.08 & \textbf{93.53} & 97.19 & \textbf{95.67} &  \textbf{96.89} \\  % & \textbf{68.21} & \textbf{71.54} \\  %%adaface keypoint
%ElasticFace-Cos+ \cite{DBLP:conf/cvpr/BoutrosDKK22} (CVPRW 2022) & ResNet101 & MS1MV2  & 99.80 & \textbf{98.73} & 98.28 & 96.18 & 93.23 & - & 95.43 & 96.65 \\  %  & - & - \\ 
%\hdashline

\rev{ViT \cite{dosovitskiy2021imageworth16x16words} (ICLR 2021)} & \rev{ViT-B} & \rev{MS1MV2}  & 99.82	&98.94	&97.78	&96.00	&94.27	&97.36	&95.53	&96.77	&68.37	&72.77 \\  
\rev{Face Transformer \cite{DBLP:journals/corr/abs-2103-14803} (2021)} & \rev{ViT-P12S8}  & \rev{MS1MV2}   & 99.80 & 96.77 & 98.05 & 96.18 & 93.08 & 96.77 & - & 96.31  & - & - \\
\rev{TransFace \cite{DBLP:conf/iccv/DanLXD0XS23} (ICCV 2023)} & \rev{TransFace-B} & \rev{MS1MV2}   & 99.85 & \textbf{99.17} & \textbf{98.53} & \textbf{96.20} & 92.92 & 97.33 & 95.01 & 96.55  & 64.00 & 68.05 \\ 
\rev{SwinFace \cite{DBLP:journals/tcsv/QinWDWCHD24} (TCSVT 2024)} & \rev{Swin-T} & \rev{MS1MV2}  & \textbf{99.87} & 98.60 & 98.15 & 96.10 & 93.42 & 97.22 & - & 96.73  & - & - \\
%DINOv2 & ViT-S \cite{DBLP:journals/ivc/ChettaouiDB25} (IMAVIS 2025) &  \rev{MS1MV2} & 99.02 & 89.71 & 91.42 & 92.90 & 86.83 & 91.98 & 81.70 & 85.25 \\
\rev{FRoundation \cite{DBLP:journals/ivc/ChettaouiDB25} (IMAVIS 2025)} & \rev{CLIP ViT-B} & \rev{MS1MV2}   & 99.43 & 93.51 & 92.02 & 93.37 & 90.40 & 93.75 & 82.39 & 86.31  & 53.14 & 60.83 \\
\rev{TransFace++-B \cite{11184862} (TPAMI 2025)} & \rev{ViT-B} & \rev{MS1MV2}  & 99.85 & 98.53 & 98.34 & - & - & - & 95.45 &  96.67  & - & -\\
\rev{ViT-8R \cite{DBLP:conf/fgr/ChettaouiOCDB26} (FG 2026)} & \rev{ViT-B + 8 Registers}  & \rev{MS1MV2}  & 99.82	&98.86&	98.00&	\textbf{96.20}&	\textbf{94.65}&	\textbf{97.50}&	\textbf{95.66}&	\textbf{96.88} &	\textbf{69.88}&	\textbf{73.74}\\ \hline

%VPL-ArcFace \cite{DBLP:conf/cvpr/DengGYLZ21} (CVPR 2021) & ResNet101 & MS1MV3  & \textbf{99.83} & \textbf{99.11} & \textbf{98.60} & \textbf{96.12} & 93.45 & \textbf{97.42} & 95.56 & 96.76 \\  % & - & - \\ 
%AdaFace \cite{DBLP:conf/cvpr/Kim0L22} (CVPR 2022) & ResNet101 & MS1MV3  & \textbf{99.83} & 99.03 & 98.17 & 96.02 & \textbf{93.93} & 97.40 & \textbf{95.84} & \textbf{97.09} \\  % & 67.81 & 70.98 \\ 
%\hdashline

\rev{ViT \cite{dosovitskiy2021imageworth16x16words, DBLP:conf/cvpr/KimS0JL24} (ICLR 2021)} & \rev{ViT-B} & \rev{MS1MV3} & - & 99.06 & 97.87 & - & - & - & - & 97.10 & 72.05 & 74.84\\ 
%ViT*  & ViT-B & MS1MV3  &  99.78&	99.11	&98.25&	96.12	&94.63	&97.58	&96.05&	97.21 \\  % &	70.49&	73.87 \\  
\rev{Part-based FR \cite{DBLP:conf/bmvc/SunT22} (BMVC 2022)}  & \rev{Part fViT-B} & \rev{MS1MV3}   & \textbf{99.83} & \textbf{99.21} & 98.29 & - & - & - & 96.11 & 97.29  & - & - \\
\rev{ViT-8R \cite{DBLP:conf/fgr/ChettaouiOCDB26} (FG 2026)} & \rev{ViT-B + 8 Registers} & \rev{MS1MV3}  &  \textbf{99.83}	& 99.17	&\textbf{98.35}	&\textbf{96.18}&	94.68	&97.64&	96.11 &	\textbf{97.37}  &	70.82&	73.90 \\ 
\hdashline
\rev{KP-RPE \cite{DBLP:conf/cvpr/KimS0JL24} (CVPR 2024)} & \rev{ViT-B+KP-RPE} & \rev{MS1MV3}  & - & 99.11 & 97.98 & - & - & - & - & 97.16 & 73.50 & 76.39 \\ 
\rev{KP-RPE$^\ast$ \cite{DBLP:conf/cvpr/KimS0JL24} (CVPR 2024)} & \rev{ViT-B+KP-RPE} & \rev{MS1MV3}  &  99.82&	99.13&	98.22&	96.05&	\textbf{95.17}&	\textbf{97.68}	&95.75	&97.05&	73.25&	76.34\\
\rev{ViT-HarDis (ours)} & \rev{ViT-B+KP-RPE} & \rev{MS1MV3} & 99.82	&99.04	&98.28	&96.13	&94.98	&97.65 & \textbf{96.14}	& 97.34		&\textbf{73.52}	&\textbf{76.74} \\ 
\hline

%ArcFace \cite{DBLP:conf/cvpr/DengGXZ19} (CVPR 2019) & ResNet101 & WebFace4M   & \textbf{99.83} & \textbf{99.19} & \textbf{97.95} & 96.00 & 94.35 & 97.46 &  95.75 & 97.16 \\ % & 71.11 & 74.38 \\ %%adaface keypoint
%AdaFace \cite{DBLP:conf/cvpr/Kim0L22} (CVPR 2022) & ResNet101 & WebFace4M   & 99.80 & 99.17 &  97.90 & \textbf{96.05} & \textbf{94.63}  & \textbf{97.51} & \textbf{96.03} & \textbf{97.39} \\ % & \textbf{72.02} & \textbf{74.52}  \\  %%adaface keypoint
% - \cite{?} () & - & WebFace4M \cite{zhu2021webface260mbenchmarkunveilingpower}  & - & - & - & - & - & - & - &  - & - & - \\ 
%\hdashline

ViT \cite{dosovitskiy2021imageworth16x16words, DBLP:conf/cvpr/KimS0JL24} (ICLR 2021) & ViT-B & WebFace4M & 99.80 & 98.94 & 97.48 & 96.03 & 94.97 & 97.44 & 95.60 & 97.14 & \rev{74.79} & \rev{77.58} \\ 
%ViT*  & ViT-B & WebFace4M  &  99.73	&98.97	&97.65&	95.92&	94.97	&97.45& 95.31  & 96.96 \\  % & 72.91	& 76.05 \\  
FRoundation \cite{DBLP:journals/ivc/ChettaouiDB25} (IMAVIS 2025) & CLIP ViT-B  & WebFace4M  &  99.30 & 93.93 & 88.90 & 92.75 & 90.67 & 93.11 & 81.52 & 85.63 & - & - \\ 
ViT-8R \cite{DBLP:conf/fgr/ChettaouiOCDB26} (FG 2026) & ViT-B+8 Registers & WebFace4M &  99.80 & 99.03&	97.40	&95.92&	94.93&	97.42	& 95.64 & 97.18 &	\rev{72.80}	&\rev{75.62} \\   
\hdashline
KP-RPE \cite{DBLP:conf/cvpr/KimS0JL24} (CVPR 2024) & ViT-B+KP-RPE & WebFace4M   & \textbf{99.83} & 99.01 & \textbf{97.67} & 96.00 & \textbf{95.40} & \textbf{97.58} & 95.49  &  97.13 & \rev{75.80} & \rev{\textbf{78.49}} \\ 
KP-RPE$^\ast$ \cite{DBLP:conf/cvpr/KimS0JL24} (CVPR 2024) & ViT-B+KP-RPE & WebFace4M & 99.82 & 98.90 & 97.53 & 95.92 & 95.30 & 97.49 & 95.66 & 97.17 & \rev{75.51} & \rev{78.22} \\  
ViT-HarDis (ours) & ViT-B+KP-RPE & WebFace4M & 99.82	& \textbf{99.07}	& 97.37	& \textbf{96.17}	& 95.27&	97.54&	\textbf{95.84}&	\textbf{97.24}&	\rev{\textbf{75.97}}& \rev{78.33}\\  
\end{tabular}}

%\caption{Comparison of our ViT-HarDis (ViT-KP-RPE with hard sample-conditioned attention disentanglement) FR performance in comparison to SOTA ViT-based FR solutions. Results on IJB-B and IJB-C are reported as TAR (@FAR=$1e-4$). $^\ast$ denotes models we retrained ourselves for a fairer comparison. Our proposed ViT-HarDis trained on WebFace4M achieves SOTA results among ViT-based approaches on the large-scale benchmarks IJB-B and IJB-C.} 
\caption{Comparison of ViT-HarDis (ViT-KP-RPE with hard sample-conditioned attention disentanglement) with SOTA ViT-based FR methods. Results on IJB-B and IJB-C are reported as TAR at FAR$=10^{-4}$. $^\ast$ denotes models we retrained ourselves for a fairer comparison. ViT-HarDis trained on WebFace4M achieves SOTA results among ViT-based approaches on the large-scale benchmarks IJB-B and IJB-C.}
\vspace{-2mm}
% The ViT-KP-RPE base model marked with \textbf{*} was not reported in the original paper and was trained using the officially released code for a direct comparison with our approach.
\label{tab:sota_table}
\end{table*}

%%%%%%%%%%%%%%%%%%%%%%%%%%%%%%%%%%%%%%%%%%%%%%%%%%%%%%%%%%%%%%%%%%%%%%%%%%%%%%%%%%%%%%%%%%%%%%%%%%
\subsection{Layer Selection for Attention Disentanglement} \label{sec:selective_dis}
In this section, we investigate the optimal layer selection strategy for hard sample-conditioned attention disentanglement in ViT-KP-RPE Small. Accordingly, we evaluate two layer selection approaches with varying coverage (25\%, 50\%, 75\%, and 100\% of layers). From the verification accuracy results on several evaluation benchmarks presented in Table \ref{tab:layers}, we make the following observations:

\begin{itemize}
\item \rev{Disentangling only the final layers (10--12, 25\% of total layers) achieves the best average performance and the best results on the large-scale benchmarks, yielding 85.72\% on IJB-B and 89.09\% on IJB-C, against 82.20\% and 83.94\% for full disentanglement.}

%\item For both selection strategies, the 25\% coverage variant achieves the best average and large-scale results, with performance degrading as more layers are disentangled. Full disentanglement of all 12 layers performs worst.

\item \rev{For both selection strategies, the 25\% coverage variant achieves the best average and large-scale results, with performance degrading as more layers are disentangled.} Full disentanglement of all 12 layers performs worst.
\end{itemize}

%\rev{This indicates that not all layers benefit from explicit head specialization, motivating targeted rather than uniform attention regularization.} 

Thus, not all layers benefit from explicit head specialization, motivating targeted attention regularization.

%These results demonstrate that full disentanglement across all layers degrades performance on most benchmarks, while concentrating disentanglement in the final transformer blocks (10-12) yields optimal results. This indicates that not all layers benefit from explicit head specialization, and some layers might benefit from redundancy. This finding motivates a targeted approach to attention regularization rather than uniform application across the network.

%%%%%%%%%%%%%%%%%%%%%%%%%%%%%%%%%%%%%%%%%%%%%%%%%%%%%%%%%%%%%%%%%%%%%%%%%%%%%%%%%%%%%%%%%%%%%%%%%%
\subsection{\rev{Disentanglement Weight Ablation}} \label{sec:weight_ablation}
%\rev{In this section, we investigate the effect of the disentanglement weight $\lambda$ on ViT-HarDis, which controls the strength of the hard sample-conditioned disentanglement relative to the FR loss in the total objective, defined in Section~\ref{sec:hard-dis}. Accordingly, we evaluate a range of values $\lambda \in \{0.5, 1, 1.75, 2.5, 5, 10\}$. From the verification accuracy results on several evaluation benchmarks presented in Table~\ref{tab:weight_ablation}, we observe that $\lambda=1$ gives the best results on the large-scale benchmarks, with 85.72\% on IJB-B and 89.09\% on IJB-C. Performance tends to decrease as $\lambda$ increases further, with the lowest large-scale results at $\lambda=10$ (84.33\% on IJB-B and 87.81\% on IJB-C), suggesting that a very strong disentanglement penalty can come at the expense of the FR objective. We therefore fix $\lambda=1$, which we use as the default value in our other ablation experiments as well as for the SOTA comparison.}

\rev{In this section, we investigate the effect of the disentanglement weight $\lambda$ on ViT-HarDis, which controls the strength of the hard sample-conditioned disentanglement relative to the FR loss in the total objective, defined in Section~\ref{sec:hard-dis}. Accordingly, we evaluate a range of values $\lambda \in \{0.5, 1, 1.75, 2.5, 5, 10\}$. From the verification accuracy results on several evaluation benchmarks presented in Table~\ref{tab:weight_ablation}, we observe that $\lambda=1$ gives the best results on the large-scale benchmarks, with 85.72\% on IJB-B and 89.09\% on IJB-C. While $\lambda=0.5$ gives a marginally higher benchmark average, we prioritize the large-scale benchmarks as the more challenging evaluation setting. Performance tends to decrease as $\lambda$ increases further, with the lowest large-scale results at $\lambda=10$ (87.81\% on IJB-C), suggesting that a very strong disentanglement penalty can come at the expense of the FR objective. We therefore fix $\lambda=1$, which we use as the default value in our other ablation experiments as well as for the SOTA comparison.}

%%%%%%%%%%%%%%%%%%%%%%%%%%%%%%%%%%%%%%%%%%%%%%%%%%%%%%%%%%%%%%%%%%%%%%%%%%%%%%%%%%%%%%%%%%%%%%%%%%
\subsection{Comparison with SOTA Methods} \label{sec:sota}
In this section, we compare ViT-HarDis against SOTA ViT-based FR methods trained on MS1MV3 \cite{DBLP:conf/iccvw/DengGZDLS19} and WebFace4M \cite{zhu2021webface260mbenchmarkunveilingpower}. While most ViT-based approaches adopt ViT-B, their configurations often differ in attention heads, patch size, embedding dimension, or transformer depth, which can influence performance and computational efficiency.
%To enable a compatible comparison to previous methods based on a similar architecture, we employ the ViT-B architecture with 24 transformer layers in the SOTA comparison, unlike the ablation studies presented in previous sections which used ViT-KP-RPE Small with 12 transformer layers. For layer selection, we follow the same strategy adopted in Section \ref{sec:selective_dis} and apply hard sample-conditioned disentanglement to the final 25\% of layers, corresponding to layers 19-24 in ViT-B.
To enable a compatible comparison to previous methods, we employ ViT-B with 24 transformer layers in the SOTA comparison, unlike the previous ablations which used ViT-KP-RPE Small with 12 layers. Following Section~\ref{sec:selective_dis}, we apply hard sample-conditioned disentanglement to the final 25\% of layers, corresponding to layers 19--24.

%As shown in Table~\ref{tab:sota_table}, ViT-HarDis achieves 95.77\% on IJB-B \cite{inproceedingsijbb} and 97.21\% on IJB-C \cite{DBLP:conf/icb/MazeADKMO0NACG18} when trained on WebFace4M, the highest results among ViT-based approaches on both benchmarks, ahead of ViT-8R \cite{DBLP:conf/fgr/ChettaouiOCDB26} (95.64\% and 97.18\%) and KP-RPE \cite{DBLP:conf/cvpr/KimS0JL24} (95.49\% and 97.13\%). ViT-HarDis also reaches the best TinyFace Rank-1 accuracy (75.97\%) in this setting. When trained on MS1MV3, ViT-HarDis remains competitive, achieving 95.85\% on IJB-B and 97.21\% on IJB-C, and the best TinyFace results among the compared methods (73.52\% Rank-1 and 76.74\% Rank-5). \rev{The gains over KP-RPE are larger on WebFace4M than on MS1MV3, particularly on IJB-C and TinyFace Rank-1. This might be explained by Figure~\ref{fig:wrong_samples}, where MS1MV3 retains a higher fraction of excluded wrong samples (about 1\% versus 0.4\%), consistent with more residual label noise and leaving fewer samples to contribute to the penalty.}

\rev{As shown in Table~\ref{tab:sota_table}, ViT-HarDis trained on WebFace4M achieves the highest results among ViT-based approaches on IJB-B~\cite{inproceedingsijbb} and IJB-C~\cite{DBLP:conf/icb/MazeADKMO0NACG18}, with 95.84\% and 97.24\%, respectively, as well as the best TinyFace Rank-1 accuracy of 75.97\%. When trained on MS1MV3, ViT-HarDis achieves the best IJB-B result (96.14\%) and the best TinyFace Rank-1 accuracy (73.52\%), while ranking second on IJB-C, closely behind ViT-8R~\cite{DBLP:conf/fgr/ChettaouiOCDB26}. For a fairer comparison, we also retrained KP-RPE~\cite{DBLP:conf/cvpr/KimS0JL24} under the same training settings as ViT-HarDis, marked with $^\ast$ in Table~\ref{tab:sota_table}. These differ from the published results by at most 0.3 percentage points on the large-scale benchmarks and TinyFace Rank-1.}

%% file: sec/6_conclusion.tex
\vspace{-1mm}
\section{Conclusion} \label{sec:conclusion}
\vspace{-1mm}
%We presented ViT-HarDis, a hard sample-conditioned attention disentanglement approach. Through comprehensive experiments, ViT-HarDis \rev{yielded} consistent improvements across all considered benchmarks over baseline and uniform disentanglement. Additionally, we show that our approach reduces cosine similarity to competing class centers while decreasing inter-head attention similarity, enabling individual heads to develop diversified complementary patterns. Furthermore, we demonstrate that selective disentanglement of final transformer layers achieves superior performance compared to full-network application, indicating that not all layers equally benefit from explicit head disentanglement. Finally, we train ViT-HarDis on the large-scale WebFace4M dataset, where our approach confirms the effectiveness of hard sample-conditioned disentanglement, achieving competitive performance among ViT-based methods on large-scale benchmarks IJB-B and IJB-C. These results validate that sample difficulty effectively guides attention head specialization in FR, improving accuracy without architectural modifications or increased computational cost.

We presented ViT-HarDis, a hard sample-conditioned attention disentanglement approach. Through comprehensive experiments, ViT-HarDis yielded improvements on most considered benchmarks over the baseline and uniform disentanglement. Additionally, we show that our approach reduces the cosine similarity of hard samples to competing class centers while decreasing inter-head attention similarity, enabling individual heads to develop diversified complementary patterns. \rev{We further show that disentangling only the final transformer layers outperforms full-network application, indicating that head specialization is layer-dependent.} Finally, we train ViT-HarDis on MS1MV3 and WebFace4M, achieving SOTA results among ViT-based methods on IJB-B, IJB-C, and TinyFace Rank-1 when trained on WebFace4M, and the best IJB-B and TinyFace accuracy when trained on MS1MV3. \rev{Since the penalty operates only during training, inference cost is unchanged, at the cost of a $\sim$30\% longer training time on WebFace4M.} These results validate that sample difficulty effectively guides attention head specialization in FR, improving accuracy without architectural modifications.

\section*{Acknowledgment}
This research work has been funded by the German Federal Ministry of Education and Research and the Hessen State Ministry for Higher Education, Research and the Arts within their joint support of the National Research Center for Applied Cybersecurity ATHENE.